\documentclass[12pt]{article}
\usepackage[utf8]{inputenc}
\usepackage[T1]{fontenc}
\usepackage{lmodern}
\usepackage[english]{babel}
\usepackage{microtype}
\usepackage{textcomp} 
\usepackage{amsmath, amssymb}
\usepackage{geometry}
\usepackage{graphicx}
\graphicspath{{./figures/}}               
\DeclareGraphicsExtensions{.pdf,.png,.jpg}

\makeatletter
\let\orig@includegraphics\includegraphics
\renewcommand{\includegraphics}[2][]{%
  \orig@includegraphics[#1]{\detokenize{#2}}%
}
\makeatother

\usepackage{booktabs}

\usepackage{amsthm}

\newtheorem{remark}{Remark}

\usepackage{hyperref}
\usepackage{framed}
\usepackage{mdframed}

\usepackage{graphicx}
\usepackage{subcaption}
\usepackage{float}
\usepackage{booktabs}
\usepackage{float}

\graphicspath{{figures/}}
\usepackage{mathtools}

\begin{document}

\title{Structural Foundations for Leading Digit Laws: Beyond Probabilistic Mixtures}
\author{
Vladimir Berman \\
Aitiologia LLC \\
\texttt{vb7654321@gmail.com}
}
\date{July 21, 2025}
\maketitle

\begin{abstract}
This article presents a modern deterministic framework for the study of leading significant digit distributions in numerical data. Rather than relying on traditional probabilistic or mixture-based explanations, we demonstrate that the observed frequencies of leading digits are determined by the underlying arithmetic, algorithmic, and structural properties of the data-generating process. Our approach centers on a shift-invariant functional equation, whose general solution is given by explicit affine-plus-periodic formulas. This structural formulation explains the diversity of digit distributions encountered in both empirical and mathematical datasets, including cases with pronounced deviations from logarithmic or scale-invariant profiles.

We systematically analyze digit distributions in finite and infinite datasets, address deterministic sequences such as prime numbers and recurrence relations, and highlight the emergence of block-structured and fractal features. The article provides critical examination of probabilistic models, explicit examples and counterexamples, and discusses limitations and open problems for further research. Overall, this work establishes a unified mathematical foundation for digital phenomena and offers a versatile toolset for modeling and analyzing digit patterns in applied and theoretical contexts.
\end{abstract}



\tableofcontents
\listoffigures

\section*{I. Introduction}
\addcontentsline{toc}{section}{I. Introduction}

The so-called "laws of digits," especially Benford's Law, have often been presented as miraculous and universal. Probabilistic literature argues that these laws emerge in sufficiently "large," "random," or "naturally occurring" datasets. A tradition that began with Newcomb and Benford, and culminated in the sophisticated mixture theories of Hill, seems to rest on the premise that the law only holds — or even makes sense — in the thermodynamic limit of infinite ensembles.

Yet this position is, upon reflection, both logically tenuous and physically suspect. First, it is unclear why a truly universal law should require a dataset to be "large enough," or why any particular threshold of data size or randomness should be privileged. Benford himself compiled and analyzed tables of modest size, and real-world datasets often contain far fewer entries than probabilistic arguments would suggest necessary. If the law of digits is genuine, its structure ought to manifest not only in immense or chaotic collections, but also in small, even singular sets, so long as their numerical content is sufficiently well-defined.

A more subtle, but deeper, issue lies in the architecture of probabilistic justifications. Most such arguments — whether invoking scale-invariance, random mixtures, or "naturalness" — are fundamentally circular. They select, often post hoc, the probability spaces or mixture distributions specifically to yield the desired law. The celebrated result of Hill, for example, asserts that if one samples from a random mixture of arbitrary distributions (with the right, carefully tuned weighting), then the outcome will almost surely satisfy Benford's Law. But this mixture is not found in nature; it is a mathematical construct, chosen precisely because it generates the law. In practical settings, random processes are governed by specific mechanisms, not by omnipresent mixtures over the space of all possible distributions. The requirement that nature itself somehow mixes all conceivable distributions in exactly the right way is not only artificial — it is implausible.

Thus, while the probabilistic paradigm has provided many beautiful results and fueled the mystique of digit laws, it has also diverted attention from the deeper, deterministic reality underlying digit phenomena. The arithmetic of digit blocks, as we demonstrate in this work, is governed by a simple, explicit recurrence that holds for any set or sequence — regardless of size, randomness, or origin. This recurrence leads directly, by analytic means, to the family of all possible digit block distributions. Benford's Law appears as a special (and important) case, but is neither unique nor universal; non-Benford distributions arise naturally and are equally valid within the same analytic framework.

Our approach, therefore, is not only conceptually distinct, but also broader and more rigorous. We show that the so-called "law of digits" is an arithmetic property, not a probabilistic one: its explanation requires no appeal to stochastic largeness, scale-invariance, or fantastical mixtures of distributions. Instead, the law follows as a deterministic consequence of the structure of number representations, valid for finite and infinite datasets alike.

This realization recasts the entire landscape of digit laws. The question is not why certain patterns appear in the data, but why the recurrence structure — simple, universal, and analytic — so robustly governs the frequencies of digit blocks across all contexts. In the chapters that follow, we develop this structure in detail, deriving both Benford and non-Benford laws as explicit solutions to a single, transparent functional equation.

\subsection*{1.2 Motivations for a Deterministic Theory}

The astonishing ubiquity of Benford’s Law has, for decades, motivated a tremendous number of probabilistic and statistical explanations—most notably those of Hill, Pinkham, and their followers. These approaches, elegant as they are, fundamentally depend on the construction of vast abstract spaces, typically assuming mixtures of distributions, scale invariance, or random sampling from infinite sets. However, this line of reasoning inevitably raises deep conceptual questions:

\begin{itemize}
    \item Why should we expect nature—or data-generating processes—to mix distributions in precisely the right (and often artificial) proportions to produce the logarithmic law?
    \item How many numbers are “enough” for the law to appear? Why should the structure fail for small or finite datasets, as is often the case in practice and even in Benford’s original empirical studies?
    \item Can a law that is only justified by the limiting behavior of enormous or abstract sets truly reflect a universal mathematical structure?
    \item Is it reasonable to believe that the emergence of digit laws in physical or social data requires fine-tuned randomness or statistical equilibrium at every scale?
\end{itemize}

The classic probabilistic framework tends to sidestep these issues, presuming that only in the “large-number limit” do the underlying mechanisms manifest. Yet, this perspective is unsatisfying from both a mathematical and a practical standpoint. In real applications—ranging from accounting and forensic science to experimental physics—we encounter datasets that are finite, often small, and far from idealized randomness.

Intuitively, if a digit law is truly fundamental, it should not be fragile: its validity should not depend on mixing “just the right” distributions, nor should it break down for small $N$. Instead, the law should emerge directly from the structure of the number system itself, with the same logic applying equally well to a single number, a handful of values, or an enormous dataset. The lack of a clear boundary between “small” and “large” $N$ in existing theories is a conceptual flaw that deterministic analysis can address.

Thus, there is a compelling need for a rigorous, constructive, and fully deterministic framework—one that
\begin{itemize}
    \item explains digit distributions from first principles, not as a statistical artifact,
    \item applies seamlessly to both finite and infinite sets,
    \item and exposes the precise mechanisms and constraints that give rise to both Benford and non-Benford laws.
\end{itemize}

In this work, we take up this challenge. We show that the structure of digit laws is dictated not by abstract probabilistic mixtures, but by explicit recurrence relations and functional equations governing the frequencies of digit blocks. Our approach reveals the mathematical “skeleton” underlying all possible digit distributions, clarifies the true scope and limitations of Benford’s Law, and eliminates the need for artificial assumptions or “randomness at infinity.”

Ultimately, this deterministic perspective provides a unified and transparent foundation for understanding digit laws—one that works in every setting, from the smallest sample to the largest dataset, and that allows us to precisely characterize the entire spectrum of admissible digit distributions.

\paragraph{Definition (Distribution of Leading Significant Digits, DLSD).}
Given a finite or infinite sequence (or set) of positive real numbers $\{x_i\}$, the \emph{Distribution of Leading Significant Digits (DLSD)} is the frequency function $\rho(k)$, $k = 1, \ldots, 9$, defined as the proportion of elements whose leading digit (in base 10) is $k$:
\[
\rho(k) = \frac{1}{N} \sum_{i=1}^{N} \mathbb{I}(d(x_i) = k),
\]
where $N$ is the total number of elements (finite or infinite), $d(x_i)$ denotes the first significant digit of $x_i$, and $\mathbb{I}$ is the indicator function.

For infinite sequences or theoretical distributions, one considers the limiting frequency as $N \to \infty$. In empirical or finite data sets, $\rho(k)$ is computed for the observed sample.


\paragraph{Distribution of leading $m$-digit blocks.}
Let $\{x_i\}_{i=1}^N$ be positive real numbers and let $m\in\mathbb{N}$.
For $x>0$, the \emph{leading $m$-digit block} is the unique integer
\[
B_m(x)\in\{10^{m-1},\dots,10^{m}-1\}
\quad\text{such that}\quad
k\,10^{n}\le x < (k+1)\,10^{n}\ \text{for some }n\in\mathbb{Z},
\]
i.e., the first $m$ significant digits of $x$ equal $k$.
The empirical frequency of block $k$ is
\[
\rho_m(k)=\frac{1}{N}\sum_{i=1}^{N}\mathbf{1}\!\bigl(B_m(x_i)=k\bigr),
\qquad k=10^{m-1},\dots,10^{m}-1,
\]
and clearly $\sum_{k=10^{m-1}}^{10^{m}-1}\rho_m(k)=1$.
The case $m=1$ gives the usual first-digit distribution $\rho_1(k)$, $k=1,\dots,9$.

\paragraph{Equivalent logarithmic check (used later).}
The condition $B_m(x)=k$ is equivalent to
\[
\log_{10}k \ \le\ \log_{10}x-n \ <\ \log_{10}(k+1)
\quad\text{for some }n\in\mathbb{Z}.
\]
(Thus one can test membership using the base-10 logarithm; a fractional-part
form will be introduced later.)

\paragraph{Benford benchmark.}
Under Benford’s law the expected frequencies for any order $m\ge1$ are
\[
\rho_m^{\mathrm{Benford}}(k)
= \log_{10}\!\left(\frac{k+1}{k}\right),
\qquad k=10^{m-1},\dots,10^{m}-1,
\]
which reduces to the standard first-digit formula when $m=1$.


\section*{II. Historical and Literature Review}
\addcontentsline{toc}{section}{II. Historical and Literature Review}

\subsection*{2.1 Early Observations: Newcomb and Benford}

The origins of digit laws can be traced back to Simon Newcomb, who, in 1881, observed that the early pages of logarithmic tables were far more worn than later ones~\cite{newcomb1881}. Although Newcomb offered only a brief empirical remark, this observation captured a subtle statistical regularity in naturally occurring numbers.

Frank Benford, in 1938, significantly expanded upon Newcomb’s idea by systematically analyzing more than 20 datasets—from river lengths to atomic weights—and famously formulated what is now called Benford’s Law~\cite{benford1938}. His empirical results, while persuasive, were still based on finite and sometimes quite modest sample sizes. The law was quickly embraced in some applied fields (such as accounting and engineering) and has inspired further compilations and reviews, and has inspired a sequence of major reviews, beginning with Raimi’s classic 1976 overview~\cite{raimi1976} and continuing in the later works of Berger and Nigrini~\cite{berger2015,nigrini2012}. For a thorough account of the mathematical theory and current perspectives on Benford’s Law, see the volume edited by Steven J. Miller~\cite{miller2015} and the foundational papers of Theodore Hill~\cite{hill1995a,hill1995b}. These works provide essential context and in-depth discussion of both classical and modern results.

\subsection*{2.2 The Search for Explanation and Early Critique}

The apparent universality of Benford’s Law soon drew mathematical attention—and skepticism.Stigler~\cite{stigler1945} provided a formal analytic approach for reconstructing the distribution of first significant digits in deterministic sequences. He demonstrated that the applicability of Benford’s Law is limited and depends on the specific structure of the underlying sequence, highlighting that many simple mathematical sequences exhibit digit distributions differing substantially from the logarithmic law. Hamming~\cite{hamming1970} and others warned that many supposed “explanations” were little more than tautologies, reflecting properties of the logarithm or number representation itself rather than genuine statistical laws.

The concept of scale-invariance became central in these debates. Pinkham~\cite{pinkham1961} elegantly demonstrated that Benford’s Law is the unique digit law invariant under scale transformations, but this invariance, while mathematically appealing, does not directly explain why such a law should arise in nature. Feller~\cite{feller1966} went further, connecting the law to the uniform distribution of mantissas (fractional parts of logarithms), but he also emphasized the limits of this approach for empirical data.

\subsection*{2.2.1 Stigler’s Contribution and the Mechanistic View of Digit Laws}

George Stigler’s 1945 note marks a fundamental turning point in the understanding of leading digit phenomena. Rather than searching for a universal law, Stigler was the first to systematically investigate how the distribution of leading digits depends on the specific mechanisms generating the data.

Stigler’s key insight was that the frequencies of first digits are not governed by a single empirical rule, but by the arithmetic and probabilistic structure of the process itself. For example, if two numbers are selected independently and uniformly from a given interval, and their product is formed, the resulting distribution of leading digits is not described by Benford’s Law, but by a distinct formula—a convolution of two uniform distributions on the logarithmic scale. Stigler showed that for each different generative process, a different digit law emerges.

Crucially, Stigler demonstrated that the celebrated logarithmic law for leading digits (now known as Benford’s Law) is not a universal phenomenon, but rather the outcome of specific multiplicative or exponential mechanisms, and only in certain limiting cases. For many other data-generating processes, especially involving sums or products of finite numbers of random variables, the digit distribution deviates from the logarithmic law.

Stigler’s mechanistic perspective thus established that the diversity of digit distributions should be understood as consequences of the particular probabilistic and arithmetic procedures underlying the data. This insight paved the way for the modern view, in which Benford’s Law is recognized as just one among many possible digit laws, each corresponding to a particular structural context.

\paragraph{A necessary critique.}
It should be noted, however, that Stigler’s analysis—while pioneering in its mechanistic perspective—also relied on the assumption that, as the process of generating numbers becomes increasingly complex or as the numbers themselves become larger, the distribution of leading digits will converge to the logarithmic law. This expectation, shared by Benford and many later authors, is not universally justified. In reality, significant deviations from Benford's Law persist even in large or complex data sets, especially when the underlying arithmetic or probabilistic structure departs from the idealized random multiplicative scenario.

Thus, both Stigler and Benford, despite their important insights, overestimated the universality of the logarithmic law in the limit. Modern analysis and explicit counterexamples show that the asymptotic regime does not guarantee convergence to Benford's Law; rather, the specific structure of data generation continues to play a critical role, regardless of scale.

\subsection*{2.3 The Era of Probabilistic Mixtures: Hill and Beyond}

The most widely cited mathematical justification came with Hill’s series of papers in the 1990s~\cite{hill1995a}. Hill showed that if one randomly samples from a sufficiently “broad” mixture of distributions, then the resulting data will, with probability one, satisfy Benford’s Law. While this theorem is technically beautiful, it rests on highly artificial and, arguably, nonphysical assumptions—namely, that the universe generates data by perfectly mixing all possible distributions in just the right way.

This “probabilistic mixture” paradigm has been both influential and controversial. Advocates point to its explanatory power and generality, while critics note that the required mixing does not correspond to any known mechanism in nature or society. Modern reviews~\cite{devlin2008, berger2015} offer both enthusiastic summaries and pointed criticisms of the mixture approach, highlighting both its strengths and conceptual limitations.

\subsection*{2.4 Deterministic and Algorithmic Perspectives}

In parallel, a line of research has sought purely deterministic and algorithmic origins for digit laws. Knuth~\cite{knuth1998} explored digit patterns in mathematical sequences and computational algorithms, showing that Benford’s Law often—but not always—appears, and that its presence depends delicately on the arithmetic structure of the process. Diaconis~\cite{diaconis1977} developed the idea of equidistribution modulo one as a tool for understanding when digit laws hold in number-theoretic sequences.

Yet, these studies, while rich in examples, ultimately reinforce the sense that Benford’s Law is far from universal: even “natural” processes can generate non-Benford distributions, and the law’s appearance (or absence) can be highly sensitive to underlying mechanisms.

\subsection*{2.5 Contemporary Directions and the Case for a Deterministic Framework}

Recent research has further broadened the scope, applying ergodic theory, group invariance, and algebraic techniques to the problem~\cite{pietronero2001, berger2015}. These works demonstrate that many different digit laws can arise from explicit construction, deterministic processes, or ergodic transformations—not just random mixtures. The literature is now replete with examples both supporting and contradicting the universalist vision, as summarized in modern surveys and expository articles.

Nevertheless, the prevailing narrative in much of the literature still leans heavily on probabilistic, often abstract, “naturalness” arguments—frequently sidestepping the deep question of whether such mechanisms exist or are even plausible in real-world settings.

\subsection*{2.6 Synthesis and Perspective}

This brief overview highlights the rich and complex history of digit laws—a story marked by empirical discoveries, elegant mathematics, and ongoing controversy over interpretation. While probabilistic mixtures, scale-invariance, and algorithmic randomness offer partial explanations, they are, at best, incomplete. The enduring puzzle remains: does Benford’s Law (and its relatives) reflect a fundamental statistical regularity, or is it a shadow cast by deeper arithmetic recurrences and analytic structures?

For the interested reader, the following works provide both entry points and deeper analysis:
\begin{itemize}
    \item \textbf{Empirical history:} Newcomb~\cite{newcomb1881}, Benford~\cite{benford1938}, Nigrini~\cite{nigrini2012}
    \item \textbf{Mathematical foundations:} Pinkham~\cite{pinkham1961}, Feller~\cite{feller1966}, Knuth~\cite{knuth1998}
    \item \textbf{Probabilistic theories:} Hill~\cite{hill1995a}, Berger~\cite{berger2015}, Devlin~\cite{devlin2008}
    \item \textbf{Algorithmic and analytic:} Diaconis~\cite{diaconis1977}, Pietronero et al.~\cite{pietronero2001}
\end{itemize}

\smallskip

In summary, while the literature is rich and diverse, the limitations of the classical approaches—especially their reliance on abstract probabilistic constructs—suggest the need for a new, fully deterministic analytic framework. This work takes up that challenge: to show, constructively, how all admissible digit laws (including Benford’s) follow directly from explicit recurrence relations and functional equations, without recourse to randomness, scale-invariance, or artificial mixtures.



\section*{III. Preliminaries and Key Definitions}
\addcontentsline{toc}{section}{III. Preliminaries and Key Definitions}

In this section, we formalize the key objects and notation used throughout the rest of the paper. Our focus is on the deterministic description of digit block frequencies for arbitrary sets or sequences of numbers, or even for a single real number’s expansion.

\subsection*{3.1 Digit Blocks, Decades, and Leading Frequency Functional}

\begin{itemize}
    \item \textbf{Digit Block ($k$):} Any finite sequence of decimal digits, considered as a positive integer. For example, $k = 1$, $k = 23$, or $k = 314$.
    \item \textbf{Decade ($d$):} The integer $d = \lfloor \log_{10} k \rfloor$, i.e., $d+1$ is the number of digits in $k$. The “$d$-th decade” refers to all numbers with $d+1$ digits.
    \item \textbf{Total Numbers in Decade ($N_d$):} $N_d = 10^{d+1} - 10^d$.
\end{itemize}

\textbf{Definition (Digit Block Frequency Functional):}  
Given a set $\mathcal{S}$ of positive integers (which may be a finite list, a sequence, or all numbers up to some limit), the frequency $\rho(k, d)$ is defined as
\[
\rho(k, d) := \frac{\#\{ n \in \mathcal{S}\ |\ n \text{ is in decade } d,\ \text{and } n \text{ starts with } k \}}{N_d}
\]
where the numerator counts how many numbers in $\mathcal{S}$, among all $d$-digit numbers, start with block $k$.

\textbf{Remark:} This definition is entirely deterministic. No assumptions of randomness, probability, or limiting behavior are required, unless otherwise specified.  
In the special case where $\mathcal{S}$ is “all $d$-digit numbers”, the denominator $N_d$ ensures that the frequencies are relative to the total possible choices in that decade.

\subsection*{3.2 The Role of Floor and Fractional Parts}

For any real $x$, we write:
\[
\lfloor x \rfloor = \text{greatest integer } \leq x,\qquad \{x\} = x - \lfloor x \rfloor
\]
so that $x = \lfloor x \rfloor + \{x\}$, with $\{x\} \in [0,1)$.

\textbf{Application:}  
For a digit block $k$, $d = \lfloor \log_{10} k \rfloor$ gives its decade, and the fractional part $\{ \log_{10} k \}$ often encodes “where” within the decade the block $k$ sits.

\subsection*{3.3 Universal Applicability: Sets, Sequences, and Single Numbers}

The above definitions apply to:
\begin{itemize}
    \item Any finite or infinite set $\mathcal{S}$ of positive integers.
    \item The sequence of terms of a mathematical or physical process.
    \item The decimal expansion of a single real number: Here, $\mathcal{S}$ consists of the “shifts” of the expansion.
\end{itemize}

\subsection*{3.4 Example Table: Block Frequency Calculation}

\begin{table}[H]
\centering
\caption{Illustration: Calculation of $\rho(k, d)$ for Selected Blocks $k$ in Decade $d = 2$ ($100 \leq n < 1000$)}
\begin{tabular}{ccc}
\toprule
Block $k$ & Numbers starting with $k$ in $[100, 999]$ & Frequency $\rho(k, 2)$ \\
\midrule
1   & 100, 101, ..., 199 & $100/900 \approx 0.111$ \\
23  & 230, 231, ..., 239 & $10/900 \approx 0.0111$ \\
314 & 314 only           & $1/900 \approx 0.0011$ \\
\bottomrule
\end{tabular}
\end{table}

\subsection*{3.5 The Cumulative Function and the Cumulative Digit Mapping (CDM)}
\label{subsec:cdf-cdm}

\paragraph{Motivation.}
Pointwise frequencies $\rho(k,d)$ are useful locally, but the global structure of digit blocks within a decade is most naturally expressed via their cumulative form. This leads to the function $w(k,d)$, a discrete analogue of a cumulative distribution function (CDF).

\paragraph{Formal definition and properties.}
Fix a decade $d$ and let $k\in[10^d,\,10^{d+1})$. Define
\begin{equation}\label{eq:def-w}
  w(k,d)\;=\;\sum_{j=10^d}^{\,k-1}\rho(j,d),
  \qquad
  w(10^d,d)=0,\quad w(10^{d+1},d)=1.
\end{equation}
Then $w(\cdot,d)$ is nondecreasing, takes values in $[0,1]$, and recovers pointwise frequencies via the discrete derivative
\begin{equation}\label{eq:rho-from-w}
  \rho(k,d)\;=\;w(k+1,d)\;-\;w(k,d).
\end{equation}

\paragraph{Cumulative Digit Mapping (CDM).}
Let $G:[0,1]\to[0,1]$ be a nondecreasing   function with $G(0)=0$ and $G(1)=1$, and let $a\in\mathbb{R}$. Define
\begin{equation}\label{eq:def-W}
  W(x) \;=\; a\,\lfloor x\rfloor \;+\; G\!\big(\{x\}\big), \qquad x\in\mathbb{R}.
\end{equation}


\noindent For $a = 1$,$k\in[10^d,10^{d+1})$, one has \textbf{two equivalent representations} for $\rho(k,d)$:
\begin{equation}\label{eq:rho-G-compact}
\begin{aligned}
\rho(k,d) &= G\!\big(\log_{10}(k+1)-d\big)\;-\;G\!\big(\log_{10}k-d\big),\\
\rho(k,d) &= \lfloor \log_{10}(k+1)\rfloor - \lfloor \log_{10}k\rfloor
            \;+\; G\!\big(\{\log_{10}(k+1)\}\big) - G\!\big(\{\log_{10}k\}\big).
\end{aligned}
\end{equation}

\noindent\textit{Validity.} The first identity holds for $10^{d}\le k \le 10^{d+1}-2$ (intra–decade).
The second holds for all $k\in[10^d,10^{d+1})$ and accounts for the decade boundary.

\noindent\textit{Validity.} 
The first equality holds for $10^{d}\le k \le 10^{d+1}-1$.
The second holds for all $k\in[10^d,10^{d+1})$.

In words, the floor part adjusts the jump at decade boundaries, while inside a decade the behavior is governed by $G$.

\paragraph{Benford as a special case.}
For $a=1$ and $G(s)=s$ we obtain $W(x)=x$ and the classical formula
\begin{equation}\label{eq:benford}
  \rho(k,d)\;=\;\log_{10}\!\Big(\frac{k+1}{k}\Big).
\end{equation}

\noindent\textit{Remark.}
The CDM formalism \eqref{eq:def-W}–\eqref{eq:rho-G-compact} applies verbatim to blocks of any length and will be used throughout the sequel.

\subsection*{3.6 Summary of Notation}
\label{subsec:summary-notation}

This subsection collects the symbols used in \S\ref{subsec:cdf-cdm} and later sections.

\begin{table}[H]
\centering
\caption{Summary of notation used for cumulative and digit–block objects.}
\begin{tabular}{ll}
\toprule
\textbf{Symbol} & \textbf{Description}\\
\midrule
$k$ & Digit block (e.g., $1$, $23$, $314$)\\
$d$ & Decade index, $d=\lfloor \log_{10}k\rfloor$; $d{+}1$ is \#digits in $k$\\
$[10^d,10^{d+1})$ & Decade-$d$ interval for integer arguments\\
$N_d$ & $10^{d+1}-10^d$\\
$\rho(k,d)$ & Relative frequency for block $k$ in decade $d$\\
$w(k,d)$ & Cumulative freq.; $\rho(k,d)=w(k+1,d)-w(k,d)$\\
$G:[0,1]\to[0,1]$ & Nondecreasing, $G(0)=0$, $G(1)=1$\\
$\lfloor x\rfloor, \{x\}$ & Floor and fractional part ($\{x\}\in[0,1)$) \\
$W(x)$ & $a\,\lfloor x\rfloor + G(\{x\})$\\
$a\in\mathbb{R}$ & Boundary jump adjuster in $W$\\
\bottomrule
\end{tabular}
\end{table}

\paragraph{Reading guide.}
The cumulative mapping $W$ links the block frequencies to the (fractional-part) driver $G$ via
\[
  \rho(k,d) \;=\; W\!\big(\log_{10}(k+1)-d\big)\;-\;W\!\big(\log_{10}k-d\big),
\]
so that inside each decade the behavior is governed by $G$, while the term $a\,\lfloor x\rfloor$ only compensates jumps at decade boundaries. The Benford case corresponds to $a=1$ and $G(s)=s$.

\subsection*{3.7 Concluding Remarks: Universality and Framework for All Future Theory}

The framework developed here is \emph{universal} and \emph{deterministic}:
- It covers arbitrary sequences, sets, or even a single real number’s decimal expansion.
- No stochastic, probabilistic, or ensemble assumptions are required—everything is constructed by counting and accumulation.
- This cumulative formalism will be essential in all that follows: the recurrences (Section IV), solution theory, and construction of new digit laws.

\paragraph{Summary.}
\begin{itemize}
    \item The cumulative function $w(k, d)$ and CDM formalism are the universal analytical tools for digit laws of arbitrary block length.
    \item All subsequent sections (IV and beyond) are built upon these foundations.
\end{itemize}




\section*{IV. Deterministic Structure and Core Equations for Digit Block Frequencies}
\addcontentsline{toc}{section}{IV. Deterministic Structure and Core Equations for Digit Block Frequencies}

\subsection*{4.1 Deterministic Definition of Leading Significant Digit Strings}

We formalize the concept of leading significant digits for any $x > 0$ as follows.  
Each number is mapped to an infinite string of decimal digits, starting with the first nonzero digit (all leading zeros and the decimal point are ignored):
\begin{itemize}
    \item $3 \equiv "3\,0\,0\,0\,\ldots"$
    \item $1.24 \equiv "1\,2\,4\,0\,0\,\ldots"$
    \item $0.0025 \equiv "2\,5\,0\,0\,\ldots"$
    \item $\pi \approx 3.1415\ldots \equiv "3\,1\,4\,1\,5\,\ldots"$
\end{itemize}
For a fixed integer $m \geq 1$, the truncation operator $D_m(x)$ extracts the first $m$ significant digits:
\[
D_m(x) = \left\lfloor 10^{m-1-\lfloor \log_{10} x \rfloor} \cdot x \right\rfloor.
\]
This definition encodes the entire DLSD framework in a deterministic way, independently of probability.

\subsection*{4.2 Fundamental Recurrence for Digit Blocks}

Let $k$ denote a digit block and $d = \lfloor \log_{10} k \rfloor$ its decade.  
For any integer $\delta > 0$, the frequency $\rho(k, d)$ satisfies the basic recurrence:
\[
    \rho(k, d) = \sum_{i=0}^{10^\delta-1} \rho(10^\delta k + i, d+\delta)
\]
This expresses that the frequency of block $k$ in decade $d$ equals the sum of the frequencies of all possible extensions of $k$ to decade $d+\delta$.

\subsection*{4.3 Cumulative Function and Functional Equation}

Define the cumulative function $w(k, d)$ by:
\[
    \rho(k, d) = w(k+1, d) - w(k, d)
\]
Plugging into the recurrence and rearranging, we have:
\[
     w(10^\delta(k+1), d+\delta) -w(k+1, d) 
    =w(10^\delta k, d+\delta) - w(k, d) 
\]
The key hypothesis is that both sides are independent of $k$:
\[
    w(10^\delta(k+1), d+\delta) - w(k+1, d) = C(\delta)
\]
where $C(\delta)$ is to be determined.

\subsection*{4.4 Logarithmic Change of Variables and the Cauchy Equation}

To motivate the algebraic structure underlying digit distributions, we briefly note that a logarithmic change of variables naturally leads to a functional equation of Cauchy type for the cumulative function. The details and the explicit solution of this equation will be developed in the following section.

\section*{4.5 Explicit Solution of the Shift Equation: Step-by-Step Construction}

\subsection*{4.5.1 Formulation of the Two Core Equations}

Consider $k$ in the range $10^n \leq k < 10^{n+1} - 1$ for some $n \geq 0$. The key shift equations are:
\begin{align}
  &w(10^\delta k, d + \delta) - w(k, d) = C(\delta), \\
  &w(10^\delta (k+1), d + \delta) - w(k+1, d) = C(\delta),
\end{align}
where $d = \lfloor \log_{10} k \rfloor$ and $\delta \in \mathbb{N}$.

\subsection*{4.5.2 Change of Variables and General Solution}

Let $x = \log_{10} k$, and define $v(x) = w(10^x, \left\lfloor x \right\rfloor)$.

The shift equation becomes:
\[
v(x+\delta) - v(x) = C(\delta).
\]


\textbf{Analysis of the Increment Function $C(\delta)$.}

To find the general solution, let us analyze the increment function $C(\delta)$. Fix any $x$ and consider two increments $\delta_1$ and $\delta_2$:
\[
\begin{aligned}
v(x+\delta_1) - v(x) &= C(\delta_1), \\
v(x+\delta_1+\delta_2) - v(x+\delta_1) &= C(\delta_2).
\end{aligned}
\]
Adding these two equations, we have:
\[
v(x+\delta_1+\delta_2) - v(x) = C(\delta_1) + C(\delta_2).
\]
But by the original shift equation, this difference must also equal $C(\delta_1 + \delta_2)$. Therefore,
\[
C(\delta_1 + \delta_2) = C(\delta_1) + C(\delta_2),
\]
which is Cauchy's functional equation for $C(\delta)$.

The general solution (for $\delta \in \mathbb{N}$) is:
\[
C(\delta) = a \delta,
\]
where $a$ is a constant (see, e.g.,~\cite{aczel1966,kuczma2009}).

Then
\[
v(x) = a\, \left\lfloor x \right\rfloor + G\left(\left\{ x \right\}\right),
\]

where $G$ is a  function and $\{x\}$ is the fractional part of $x$.

\subsection*{4.5.3 The Case $k = 10^n - 1$: Boundary Solution}

At the block boundary, $k = 10^n - 1$, so $x = n$ is integer.  
Let $P(n) = w(10^n, n-1)$, then the recurrence for integer arguments is:
\[
P(n+\delta) - P(n) = a \delta,
\]
with general solution
\[
P(n) = a n + \text{const}.
\]

Thus, at the boundaries, the cumulative function is affine in $n$, matching the piecewise structure defined above.

\subsection*{4.5.4 Parallel Construction for $k+1$}

For $k+1$, let $x' = \log_{10}(k+1)$ and $v'(x') = w(10^{x'}, \left\lfloor x' \right\rfloor)$. The same reasoning gives:
\[
v'(x') = a\, \left\lfloor x' \right\rfloor + G'\left( \left\{ x' \right\} \right).
\]

\subsection*{4.5.5 Equivalence with a two–term shift form and an
asymptotic expansion}\label{subsec:two-term-proof}
Recall the four–term expression obtained in~\S4.5.5:
\[
   \rho(k,d)
   = a\!\left(\Bigl\lfloor \log_{10}(k+1)\Bigr\rfloor
             -\Bigl\lfloor \log_{10}k\Bigr\rfloor\right)
     + G\!\bigl(\{\log_{10}(k+1)\}\bigr)
     - G\!\bigl(\{\log_{10}k\}\bigr),
   \tag{4.18}
\]
where \(G\colon[0,1]\to\mathbb R\) , \(G(0)=0\) and
\(G(1)=1\).

\paragraph{Alternative two–term representation.}
For any real shift \(d\in[0,1)\) define
\[
   \rho^\ast(k,d)
   \;=\;
   G\!\bigl(\log_{10}(k+1)-d\bigr)
   -G\!\bigl(\log_{10}k-d\bigr).
   \tag{4.19}
\]
We show that \(\rho^\ast(k,d)\equiv\rho(k,d)\) for all \(k\ge 1\)
and every digit-block length \(d\).

\begin{proof}[Proof of equivalence]
Write \(x=\log_{10}k\) and \(\Delta=\log_{10}(k+1)-\log_{10}k
       =\log_{10}(1+1/k)\).
Set \(s=\{x-d\}\) and note that
\(\{x\}=s+d\pmod 1\), \(\{x+\Delta-d\}=s+\Delta\pmod 1\).
Two cases:

\smallskip
\emph{1.~Interior blocks \((d\notin\mathbb Z)\).}  
Because \(0<d<1\), the integer parts satisfy
\(\lfloor x+\Delta\rfloor=\lfloor x\rfloor\); hence the
first term in~(4.18) vanishes.  We obtain
\[
   \rho(k,d)=G(s+\Delta)-G(s)=\rho^\ast(k,d).
\]

\smallskip
\emph{2.~Boundary blocks \((d=n\in\mathbb Z_{\ge 0})\).}  
Now \(x-n=\log_{10}k-n\) has fractional part \(\{x\}\) and
integer part \(\lfloor x\rfloor-n\).  A direct calculation gives
\[
   \rho^\ast(k,n)
   = a(\lfloor x+\Delta\rfloor-\lfloor x\rfloor)
     +G(\{x+\Delta\})-G(\{x\})
   =\rho(k,n),
\]
because subtracting an integer shift \(n\) changes neither the
fractional parts nor the difference of integer parts.

Thus (4.18) and (4.19) coincide in all cases.
\end{proof}

\paragraph{Large–\(k\) asymptotics.}
Formula~(4.19) is convenient for expansions, because
\(\Delta=\log_{10}(1+1/k)=\dfrac{1}{k\ln 10}+O(k^{-2})\).
If \(G\) is differentiable at \(s=\{\log_{10}k-d\}\), a first-order
Taylor expansion yields
\[
\boxed{\;
   \rho(k,d)
   = \frac{G^{\prime}\!\bigl(\{\log_{10}k-d\}\bigr)}{k\ln 10}
     +O\!\left(\frac{1}{k^{2}}\right)
   }\qquad(k\to\infty).
\]
 This makes~(4.19) the natural starting point for estimating convergence
rates or constructing confidence bands for empirical leading-digit
frequencies.

\subsection*{4.5.6 Summary}

\begin{itemize}
    \item The general solution is $a\, \left\lfloor x \right\rfloor + G\left( \left\{ x \right\} \right)$ for all $k$ not at the boundary, and $P(n) = a n + \text{const}$ for integer boundary points.
    \item All recurrence conditions are satisfied, and the solutions coincide at the decade transitions.
\end{itemize}

\textbf{Remark:}
This construction gives the most general form of the admissible cumulative functions for DLSDs, as required by the shift recurrence. The solution directly links the theory to the structure of the Cauchy equation and makes explicit the piecewise-logarithmic nature of digital laws.

\begin{framed}
\noindent
\textbf{Main formula for the distribution of leading digits:}
\[
\rho(k, d) = a \left( \left\lfloor \log_{10}(k+1) \right\rfloor - \left\lfloor \log_{10} k \right\rfloor \right)
+ G\left( \left\{ \log_{10}(k+1) \right\} \right) - G\left( \left\{ \log_{10} k \right\} \right),
\]
where $a$ is a constant, $G$ is a  function, and $\left\{\,x\,\right\}$ denotes the fractional part of $x$.

By normalization, $G$ can be chosen so that $G(0)=0$, $G(1)=1$.
\end{framed}

\subsection*{4.6 Boundary Cases and Interpretation}

At the upper boundary of each decade, that is for \( k = 10^{d+1} - 1 \), the formula for the digit block frequency simplifies due to the properties of the function \( G \):

\[
\sum_{k=10^{d}}^{10^{d+1}-1} \rho(k,d)
  \;=\; G(1) - G(0) \;=\; 1 .
\]
We impose the normalization conditions on \( G \) to guarantee the correct summation over all possible blocks within a decade. Specifically, we require

\[
G(0) = 0,\quad G(1) = 1.
\]

 The function \( G \) can be any non-decreasing function on \([0, 1]\) with these endpoint values, but the structure of the solution for \( \rho \) is completely determined by this form, regardless of whether there are finitely or infinitely many possible solutions for \( G \). No other structural forms for \( \rho \) are possible within this framework.

\vspace{0.2cm}
\noindent
\textbf{Definition: Leading Digit Profile Function (LDPF).}

Whenever we refer to the function $G(s)$, which encodes the cumulative distribution of leading digits for a given probability density $f(x)$, we will call it the \textbf{Leading Digit Profile Function (LDPF)}. Explicitly,
\[
\mathrm{LDPF}(s) := G(s) = \int_{A}^{B} f(x)\, M(s, x)\, dx,
\]

where $M(s, x)$ is a measurable indicator function that counts whether $x$ contributes to the leading digit profile up to $10^s$. The explicit form of $M(s, x)$ will be developed in Section~V.

\textit{Throughout this work, we will use the abbreviation LDPF for clarity and consistency.}

\subsection*{4.7 Remarks}

\begin{figure}[h]
  \centering
  \includegraphics[width=0.72\linewidth]{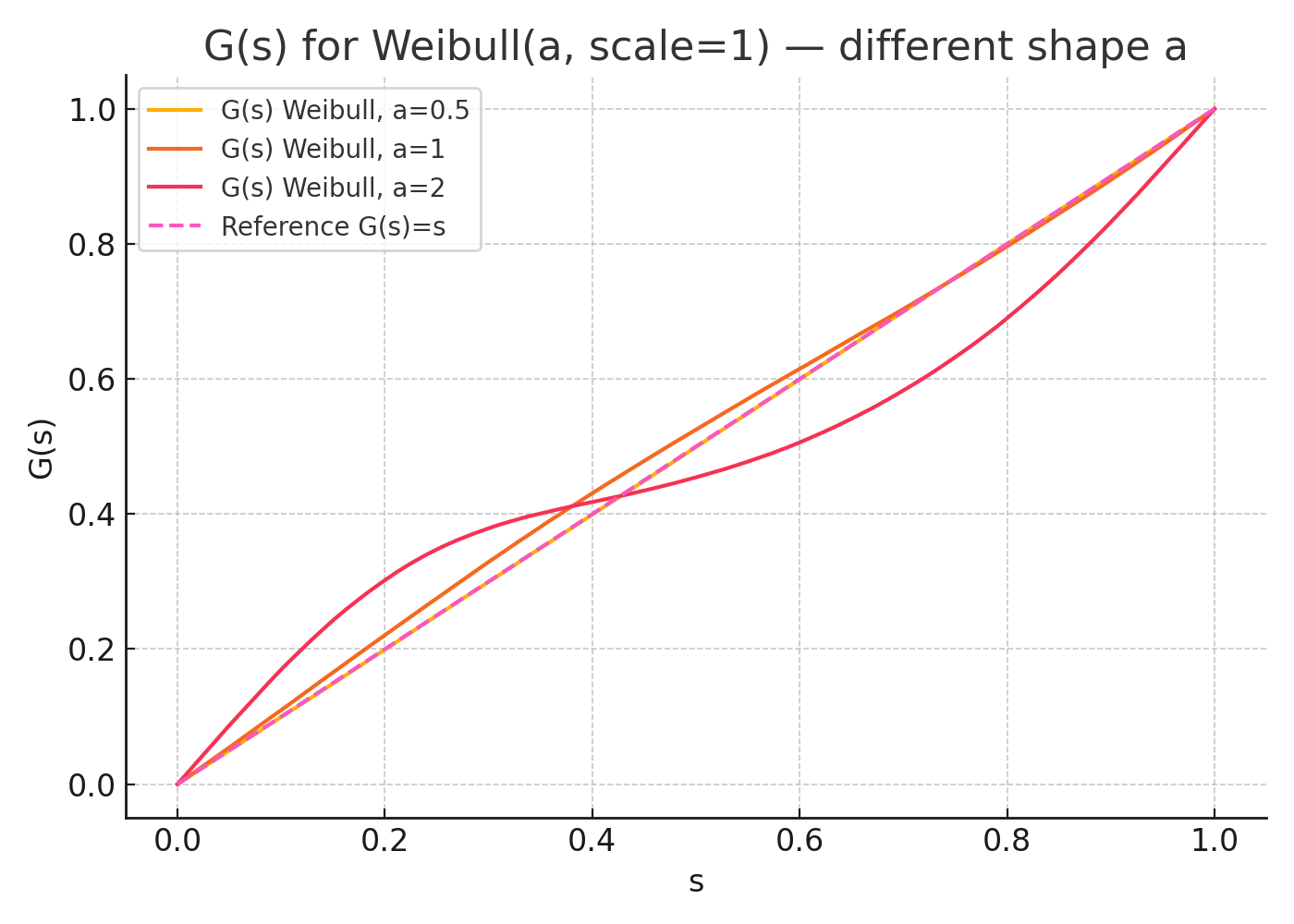}
 \caption{Empirical windowed profiles $G(s)$ for Weibull$(a,b=1)$
with shapes $a\in\{0.5,1,2\}$. Each curve is based on $N=600{,}000$ samples.
The dashed line shows the reference $G(s)=s$.}

  \label{fig:weibull_Gs_shapes}
\end{figure}

\begin{itemize}
    \item $G(s)\equiv s$ gives Benford; arbitrary $G$ gives all admissible deterministic DLSDs.
    \item All formulae and frequencies derive directly from deterministic structure, with no stochastic or “random mixture” hypotheses.

\subsection*{4.8 Deterministic Profiles versus Hill’s Random–Mixture Theorem}
Hill’s seminal 1995 theorem states that if one draws a positive‐valued random variable $X$ by \emph{first} sampling a distribution $F$ at random from a sufficiently broad, parameter–free ensemble $\mathcal F$ and \emph{then} sampling $X\sim F$, the leading‐digit law of the mixture converges in probability to Benford’s vector
\[
\Pr\bigl(\operatorname{LD}(X)=d\bigr)=\log_{10}\!\bigl(1+\tfrac1d\bigr),\qquad d=1,\dots,9.
\]
\[
\mathrm{LD}(x)=\big\lfloor 10^{\{\log_{10} x\}}\big\rfloor,\quad x>0.
\]

This conclusion hinges on two probabilistic layers—(i) an \emph{external} randomness in the choice of $F\in\mathcal F$, and (ii) the ordinary sampling randomness of $X\mid F$.

By contrast, the deterministic framework developed here requires no external mixing: a single distribution with cumulative function $F(x)$ generates a \emph{windowed profile}
\[
G(s)\;=\;\frac{1}{F(B)-F(A)}\sum_{i=m}^{n-1}\bigl[F(10^{s+i})-F(10^{i})\bigr],
\qquad s\in[0,1),\; A=10^{m},\,B=10^{n},
\]
whose analytic form $F_{W}(s;b,a)$ is governed by the shape parameters $(a,b)$ of Definition 4.2.  
When $(a,b)=(1,10)$ the profile collapses to the diagonal $G(s)=s$ and Benford’s leading‐digit frequencies reappear, but for $a\neq1$ the profile deviates in a predictable, \emph{non‐random} way—e.g.\ an $S$–shape for $a>1$ and a concave bend for $0<a<1$ (see Fig.~\ref{fig:weibull_Gs_shapes}).  Thus Hill’s law emerges here as a singular limiting case of a broader structural family, not an inevitable consequence of “generic” randomness. 

%
Empirically, 2023 country populations (UN WPP 2024; \cite{UN_WPP2024_TotalPop_Var12})
are well approximated by a Weibull model with fitted shape $\hat a \approx 0.47$.

Under the Weibull model, the shape parameter $a$ governs departures from Benford:
for small $a$ (e.g., $a\approx 0.47$) the windowed profile $G(s)$ lies close to the
Benford line $G(s)=s$, whereas as $a$ approaches and exceeds $1$ the curve becomes
pronouncedly S–shaped and the resulting leading–digit frequencies deviate
substantially from Benford’s law.

\end{itemize}




\section*{V. Euler--Maclaurin Framework}\label{sec:euler-maclaurin}
\addcontentsline{toc}{section}{V. Euler--Maclaurin Framework}

\subsection*{5.1 Motivation and Historical Context}

The Euler–Maclaurin summation formula, developed in the 18th century by Leonhard Euler and Colin Maclaurin, plays a foundational role in analytic number theory and the modern analysis of digital statistics.  
Contrary to common belief, this formula is \emph{not merely an asymptotic or approximate tool}, but rather a strictly \emph{exact decomposition} for sums of sufficiently regular functions over integers.

\subsection*{5.2 The Exact Formula and Its Terms}

Let $f(x)$ be a function defined on the real line, and let $a, b \in \mathbb{Z}$ with $a < b$.  
	\emph{The Euler–Maclaurin formula states:}
\[
\sum_{n=a}^{b} f(n) = \int_{a}^{b} f(x)\,dx
+ \frac{1}{2}\big[ f(a) + f(b) \big]
+ \int_{a}^{b} \left( x - \lfloor x \rfloor - \frac{1}{2} \right) f'(x)\,dx
\]
\textbf{Each term has a clear interpretation:}
\begin{enumerate}
    \item The main integral: $\int_{a}^{b} f(x)\,dx$ is the “continuous” contribution,
    \item The boundary average: $\frac{1}{2}[f(a) + f(b)]$ corrects for discrete endpoints,
    \item The \emph{periodic correction}: $\int_{a}^{b} (x-\lfloor x\rfloor-1/2)f'(x)\,dx$ accounts for the “fine structure” lost when summing only over integers.
\end{enumerate}

\paragraph{Key property:}  
\textit{If $f(x)$ is extended to be zero outside $[a,b]$, the above formula remains strictly valid (all other integrals vanish).  
In DLSD and related problems, this extension is natural and universal.}

\subsection*{5.3 Why Euler–Maclaurin Is Not an Approximation}

It is essential to emphasize:  
\textbf{The Euler–Maclaurin formula is exact, not just asymptotic!}  
The error arises only if one truncates after the first terms of the more general expansion (involving higher Bernoulli numbers and derivatives).  
In the version above, if $f$ is continuously differentiable and of finite support, the formula holds with perfect equality.

\subsection*{5.4 Application to Digital Block Statistics and DLSD}

This exact bridge is central for digital block statistics:  
It enables the analytic transformation of any sum over digits or blocks
\[
S = \sum_{k=a}^{b} f(k)
\]
to its integral analog, with precise control of the correction term.  
For digital statistics, $f(k)$ is often the block frequency or its cumulative, and the periodic term encodes fluctuations due to the discreteness of digits.

\subsection*{5.5 Extension to Infinite Intervals}

In applications, one often needs to sum over all possible decades (e.g., $k \in [10^{d}, 10^{d+1})$ for all $d$).  
By extending $f(x) \equiv 0$ outside its natural domain, all integrals and sums can be taken over the full real line, reducing technical complications.

\subsection*{5.6 Towards a Generalized Cumulative Function}

The preceding analysis sets the stage for introducing a generalized cumulative function, denoted here as $G(s)$, which fully characterizes the family of admissible distributions of leading significant digits (DLSDs). In the following sections, we develop the machinery necessary to construct explicit solutions for $G(s)$, and demonstrate how this approach reveals the underlying deterministic structure of digital laws.

\subsection*{5.7 The "Integral Replacement" Fallacy in the Literature}

It is common in some recent literature to claim a "proof" of Benford's law by simply replacing the discrete sum over blocks $k$ with a continuous integral over $x$.  
In this approach, the Euler–Maclaurin formula is (sometimes implicitly) invoked, but only the main integral term is retained, while both the boundary and periodic correction terms are neglected:
\[
\sum_{k} f(k) \overset{?}{\approx} \int f(x)\,dx
\]
This is often justified by the claim that, if $f$ is a probability density function (PDF) on $(0, \infty)$, the other terms are negligible, and the Benford distribution emerges directly from integrating the logarithmic PDF over the leading digit blocks.

\textbf{However, this replacement is logically flawed in the context of DLSD analysis:}
\begin{itemize}
    \item The boundary term and, more importantly, the periodic correction term in the Euler–Maclaurin formula are not in general negligible.
    \item In fact, these terms are \emph{responsible for all possible deviations from the Benford law}; neglecting them is tantamount to presupposing the result one wishes to derive.
    \item In the exact Euler–Maclaurin framework, if these terms are omitted, the remaining integral gives only a trivial result or the Benford law, regardless of the true structure of $f$.
\end{itemize}

This logical sleight-of-hand leads to circular reasoning:  
First, one assumes that a sum can be replaced by an integral (which is only strictly valid in special cases), and then one “derives” the Benford law, not noticing that all non-Benford effects have been discarded by construction.

See ~\cite{berger2015} for a critical discussion. In contrast, ~\cite{su2014} is a notable example of the \emph{incorrect} application of this principle: the authors adopt the condition as a natural assumption without any critical examination. Our analysis demonstrates that this uncritical acceptance leads to a circular justification and ultimately undermines the logical basis of the argument.

\subsubsection*{Key Point}
The Euler–Maclaurin formula shows that the Benford law arises \emph{only if} the correction terms vanish or cancel. In general, for arbitrary distributions, the full formula—including periodic corrections—must be used to capture the true digit law, and these corrections can be substantial.

\section*{5.8 Connection of the Summation Formula with the Euler--Maclaurin (EM) Representation}

Consider the sum
\[
S = \sum_{i=-\infty}^{\infty} \int_{k 10^i}^{(k+1)10^i} f(x)\,dx,
\]
where \( f(x) \) is a non-negative PDF supported on \( (0, \infty) \), and \( k \) is a fixed positive integer block.

Applying the Euler–Maclaurin summation formula, the sum can be decomposed as
\[
S = J_1 + J_2 + J_3,
\]
where:
\begin{align*}
J_1 &= \int_{-\infty}^{\infty} \left[ F((k+1)10^y) - F(k10^y) \right] dy,\\
J_2 &= 0,\\
J_3 &= \int_{-\infty}^{\infty} \frac{\partial}{\partial y} \left[ F((k+1)10^y) - F(k10^y) \right] \cdot (y - \lfloor y \rfloor - \tfrac{1}{2})\, dy.
\end{align*}
Here, \( F(x) \) is the cumulative distribution function (CDF) of \( f(x) \).

\subsubsection*{5.8.1 Evaluation of \( J_1 \)}

Make the substitution \( x = (k+1)10^y \) and \( x = k10^y \) in \( J_1 \), which yields:
\[
J_1 = \int_{0}^{\infty} \left[ \log_{10}\left( \frac{x}{k} \right) - \log_{10}\left( \frac{x}{k+1} \right) \right] f(x) dx.
\]
This further simplifies to:
\[
J_1 = \left( \log_{10}(k+1) - \log_{10}(k) \right) \int_{0}^{\infty} f(x)\,dx,
\]
which, for a properly normalized PDF, gives the Benford probability for block \( k \).

\subsubsection*{5.8.2 Evaluation of \( J_2 \)}

This term vanishes due to the integration over the full real line and the properties of the PDF:
\[
J_2 = 0.
\]

\subsubsection*{5.8.3 Structure and Role of \( J_3 \)}

The term \( J_3 \) is
\[
J_3 = \int_{-\infty}^{\infty} \frac{\partial}{\partial y} \left[ F((k+1)10^y) - F(k10^y) \right] \cdot (y - \lfloor y \rfloor - \tfrac{1}{2})\, dy.
\]
Note that, after a variable change and analysis (see detailed derivation above), \emph{the first summand in \( J_3 \) exactly cancels \( -J_1 \)}, so that no trace of the Benford law remains here.

The condition for the sum to yield Benford's law is that the \emph{entire \( J_3 \) vanishes}. This is a highly special (exotic) situation: generally, \( J_3 \neq 0 \), and the deviation from Benford is encoded precisely in \( J_3 \).

In the general case, the term \( J_3 \) can be written as:
\[
J_3 = \int_{0}^{\infty} f(x)\, V(k,x)\,dx- J_1,
\]
where
\[
V(k,x) =\left \lfloor \log_{10}\left( \frac{x}{k} \right) \right\rfloor - \left\lfloor \log_{10}\left( \frac{x}{k+1} \right) \right\rfloor,
\]
or, equivalently,
\[
V(k,x) = \lfloor \{ \log_{10}(x) \} - \{ \log_{10}(k) \} \rfloor - \lfloor \{ \log_{10}(x) \} - \{ \log_{10}(k+1) \} \rfloor + \lfloor \log_{10}(k+1) \rfloor - \lfloor \log_{10}(k) \rfloor.
\]

\textbf{Conclusion:}  
The Benford law arises if and only if the sum of all nontrivial terms \( J_3 \) vanishes for all \( k \), i.e., when the additional contributions from the oscillatory kernel \( V(k,x) \) integrate to zero with respect to \( f(x) \).

\textbf{Connection to \( \rho(k, d) \):}  
In this analysis, the sum \( S \) (and the EM decomposition) provides a direct integral representation for the block frequency \( \rho(k, d) \), and reveals exactly how deviations from Benford's law are encoded by the structure of the PDF and the periodic structure of the EM formula.


\begin{center}
\setlength{\fboxsep}{10pt}
\noindent
\framebox{
  \parbox{0.95\textwidth}{
    \textbf{Key Formula:} For any probability density $f(x)$ on $\mathbb{R}_+$, the relative frequency of leading digit block $k$ (in its corresponding decade) is given by
    \[
      \boxed{
      \rho\bigl(k,\,\lfloor\log_{10}k\rfloor\bigr) = \int_0^{\infty} f(x)\, V(k,x)\, dx
      }
    \]
    where
    \[
      V(k,x) = \lfloor \log_{10}(x/k) \rfloor - \lfloor \log_{10}(x/(k+1)) \rfloor.
    \]
  }
}
\end{center}

\section*{5.9 Connection Between the Integral Representation and $G(s)$ Formulation}

\vspace{0.5em}

From the integral (Euler-Maclaurin, EM) approach, for any $s \in [0,1)$, we define:
\begin{framed}
\[
G(s) = \int_0^\infty f(x)\, M(s, x)\, dx
\]
where
\[
M(s, x) = - \lfloor \{ \log_{10} x \} - s  \rfloor
\]
and $\{\cdot\}$ denotes the fractional part.
\end{framed}

This establishes a direct and explicit link between the analytic (integral) and the functional ($G(s)$) characterizations of digit laws.

\vspace{0.5em}

More generally, for $d = \lfloor \log_{10} k \rfloor$, we may write:
\[
\rho(k,\, \lfloor \log_{10} k \rfloor) = \left\lfloor \log_{10}(k+1) \right\rfloor - \left\lfloor \log_{10}k \right\rfloor + G(\{\log_{10}(k+1)\}) - G(\{\log_{10}k\}),
\]
where $\{\cdot\}$ denotes the fractional part. This formula separates the “main” interval crossing (as $k$ passes a new decade) from the finer structure controlled by the continuous $G(s)$.

\begin{framed}
\textbf{Key Formula:}
\[
\rho(k,\, \lfloor \log_{10} k \rfloor) = \left\lfloor \log_{10}(k+1) \right\rfloor - \left\lfloor \log_{10}k \right\rfloor + G(\{\log_{10}(k+1)\}) - G(\{\log_{10}k\})
\]
with
\[
G(s) = \int_0^\infty f(x)\, M(s, x)\, dx, \qquad M(s, x) = -\left\lfloor \{ \log_{10} x \} - s \right\rfloor
\]

\end{framed}

\section*{VI. Examples of Continuous Distributions}

\addcontentsline{toc}{section}{VI. Examples of Continuous Distributions}

\section*{6.1 Distribution Examples and Leading Digit Analysis}
\label{sec:distribution_examples}

\paragraph{Windowed digital profile.}
In practice we observe a variable $X$ only on a finite magnitude window $[A,B]$. Throughout this section we therefore study the \emph{windowed} profile $G_{[A,B]}(s)$—the cumulative distribution of the phases $\{\log_{10}X\}$ restricted to $[A,B]$ and renormalized—so that comparisons across datasets and limits as $A\downarrow 0$, $B\uparrow\infty$ are meaningful.

Let $f(x)$ be a probability density function (PDF) supported on an interval $[A, B] \subset \mathbb{R}_+$, where $0 \leq A < B \leq \infty$. The key functional object in digital leading significant digit (DLSD) analysis is the cumulative digit profile $G(s)$, defined as
\[
G(s) = \int_A^B f(x)\, M(s, x)\, dx, \qquad M(s, x) = -\left\{\log_{10} x - s \right\},
\]
where $\{\cdot\}$ denotes the fractional part. This formula connects any continuous distribution to its digital signature through logarithmic folding.

\subsection*{6.1.1 Logarithmic Partitioning: Summation Formula for $g(s)$ and $G(S)$}

To further clarify the construction, we define the function $g(s)$ as a sum over logarithmically partitioned intervals within the range $[A, B]$:
\[
g(s) = \sum_{i = \lceil \log_{10}(A) \rceil}^{\lfloor \log_{10}(B) \rfloor}
\begin{cases}
F\left(\min(10^{i+s}, B)\right) - F\left(\max(10^i, A)\right) & \text{if } [10^i, 10^{i+s}) \subseteq [A, B], \\
0 & \text{otherwise}
\end{cases}
\]
The normalized version is then
\[
G(S) = \frac{g(s)}{F(B) - F(A)}
\]
where $F$ is an arbitrary function (typically the CDF of the underlying distribution), and the summation considers only those intervals $[10^i, 10^{i+s})$ that are fully contained within $[A, B]$. The normalization ensures that $G(S)$ represents the fraction of the total change in $F$ over the interval $[A, B]$ contributed by these log-partitioned intervals.

\subsection*{6.1.2 Sampling and Leading Digit Extraction in $[1, 10)$: Illustrative Example and Justification}

A key motivation for focusing on intervals like $[1, 10)$ (or, more generally, intervals that are integer multiples of a decade such as $[10^{-2}, 10^{-1})$, $[10^k, 10^{k+1})$) lies in their remarkable analytical simplicity and direct relationship between the cumulative distribution function (CDF) of the underlying variable and the leading digit profile function $G(s)$.

To see why this is so, observe that for such intervals, the mapping from the original probability distribution $f(x)$ to the digital profile $G(s)$ simplifies dramatically. Specifically, when considering a probability density $f(x)$ supported on $[A, B]$ where $A$ and $B$ are integer powers of ten, the cumulative digit profile $G(s)$ is given by the simple normalized difference of the CDF:
\[
G(s) = \frac{F(10^s) - F(A)}{F(B) - F(A)},
\]
where $F(x)$ is the CDF of $f(x)$. In particular, for the standard decade $[1,10)$, this formula reduces to:
\[
G(s) = F(10^s) - F(1),
\]
assuming normalization over the decade.

This case is not only mathematically elegant but also serves as the most transparent illustration of the deterministic link between the structure of the underlying distribution and the observed leading digit frequencies. The simplicity of the formula for $G(s)$ in these intervals allows one to immediately "see" the digital signature of any distribution.

\textbf{Why focus on such intervals?}  
Most real-world datasets, and almost all practical illustrations of Benford-like phenomena, involve data spanning several orders of magnitude. By focusing on samples extracted from full decades, we ensure that every possible leading digit (from 1 to 9) is represented, and the counting of their frequencies becomes not only meaningful but also analytically tractable.

Moreover, in numerical experiments, it is common to start with a random sample from an underlying distribution on $(0, \infty)$ and then restrict attention to those values that fall within a chosen decade, e.g., $[1, 10)$, $[10, 100)$, etc. By examining the empirical frequencies of leading digits in these intervals, we obtain a direct and visually intuitive illustration of how the theoretical digital profile $G(s)$ manifests in practice.

\textbf{Generalization: Multiple Decades as Approximation}  
It is also possible, and often useful, to consider not just a single decade, but a union of several adjacent decades, for example, the interval $[10^{-3}, 10^{5})$. In practice, analyzing several consecutive decades provides an even better approximation to the full distribution of leading digits over the positive real axis. Remarkably, even a single decade (e.g., $[1, 10)$) can serve as a good approximation, especially when the underlying data are broadly distributed. When considering multiple decades, the cumulative contribution of all leading digits across these intervals increasingly reflects the global structure of the distribution. Thus, the choice of interval width and position becomes a matter of approximation quality and practical convenience.

\textbf{Example:}  
Suppose we sample a large number of values from an underlying distribution $f(x)$ on $(0, \infty)$, but then only analyze those that fall in $[1, 10)$, or in a wider interval such as $[10^{-3}, 10^{5})$. For each value, we record the leading digit $d \in \{1,2,\ldots,9\}$ and plot the empirical frequencies. This simple experiment immediately reveals the structure imposed by $G(s)$: for the uniform distribution, $G(s)$ is linear; for other distributions, the shape of $G(s)$ directly dictates the observed digit frequencies. With multiple decades, the observed frequencies converge ever closer to the theoretical digital profile dictated by the underlying distribution.

\textbf{Summary:}  
Intervals of the form $[1, 10)$, or more generally $[10^a, 10^b)$, are not only convenient but also fundamental for both theoretical analysis and practical illustration of leading digit laws. They embody the most direct link between the analytic form of the underlying distribution and the digital structure captured by $G(s)$. Considering multiple consecutive decades within the positive real axis (e.g., $[10^{a}, 10^{b})$ for some integers $a < b$) yields a digital profile that closely approximates the theoretical distribution over all positive numbers. This offers a practical bridge between what is observable in finite samples and the limiting behavior of leading digits as the sampled range grows to encompass the entire positive half-line.

\subsubsection*{6.1.2.1 Digital Profiles for Scale-Invariant Densities over Multiple Decades}

Consider the class of scale-invariant densities of the form:
\[
f(x) = \frac{\psi(\{\log_{10} x\})}{x},
\]
where $\psi$ is any $1$-periodic function, $\psi(t) = \psi(t+1)$, and $\{y\}$ denotes the fractional part of $y$.

Suppose we restrict attention to a union of $N$ consecutive decades, i.e., the interval $[10^{m_0}, 10^{m_1})$ with $m_0, m_1 \in \mathbb{Z}$, $N = m_1 - m_0 > 0$. Let $s \in (0, 1)$, and let $M(s, x) = 1$ if $\log_{10} x \bmod 1 < s$, and $M(s, x) = 0$ otherwise.

The cumulative digital profile function is then given by
\[
G(s) = \int_{10^{m_0}}^{10^{m_1}} f(x) M(s, x) dx = \int_{10^{m_0}}^{10^{m_1}} \frac{\psi(\{\log_{10} x\})}{x} M(s, x)\, dx.
\]

Applying the substitution $t = \log_{10} x$, $x = 10^t$, $dx = 10^t \ln 10\, dt$, and $dx/x = \ln 10\, dt$, we rewrite the integral as
\[
G(s) = \ln 10 \int_{m_0}^{m_1} \psi(\{t\}) \cdot \mathbf{1}_{\{\{t\} < s\}}\, dt,
\]
where $\mathbf{1}_{\{\{t\} < s\}}$ is the indicator function, equal to 1 if the fractional part of $t$ is less than $s$.

Since $\psi$ is $1$-periodic, we can partition the integration interval into $N$ full decades:
\[
G(s) = \ln 10 \sum_{k=0}^{N-1} \int_{0}^{1} \psi(\tau)\, \mathbf{1}_{\{\tau < s\}} d\tau
= N \ln 10 \int_0^s \psi(\tau)\, d\tau,
\]
where we set $t = m_0 + k + \tau$, $\tau \in [0,1)$.

\textbf{Final explicit formula:}
\[
\boxed{
G(s) = (m_1 - m_0)\, \ln 10 \int_0^s \psi(\tau)\, d\tau, \qquad 0 < s < 1
}
\]
This formula shows that for any scale-invariant density of this form, the cumulative digital profile over $N$ consecutive decades is completely determined by the integral of the generating function $\psi$ up to $s$, multiplied by the number of decades and the constant $\ln 10$.

\textbf{Special case: Benford's Law.} For $\psi(\tau) \equiv 1/((m_1 - m_0)\, \ln 10)$ (i.e., $f(x) = 1/(x\ln 10)$), this reduces to the linear profile:
\[
G(s) = s.
\]

\textbf{Remark.}  
This result remains valid if one considers any collection of $N$ separate, non-overlapping decades; the only requirement is that the union of all considered intervals has total logarithmic width $N$.


\subsubsection*{6.1.2.2 Multi-Decade Windows and Approximation to the Global Profile}

Let \(F\) be the CDF of the underlying density on \((0,\infty)\). For an integer window
of \(N\) consecutive decades \([10^{m_0},\,10^{m_0+N})\), define the normalized cumulative
digital profile
\[
G_N(s)
=
\frac{\displaystyle \sum_{i=m_0}^{m_0+N-1} \bigl[F(10^{\,i+s})-F(10^{\,i})\bigr]}
     {\displaystyle F(10^{\,m_0+N})-F(10^{\,m_0})}, \qquad s\in[0,1].
\]
This quantity aggregates the contribution of each decade in the window and
renormalizes by the total mass in the window. In practice, even a single decade
(\(N=1\), e.g. \([1,10)\)) can be a serviceable approximation when most probability
mass lies in that decade; including additional adjacent decades typically yields a
markedly closer match to the global digital profile.

\begin{figure}[t]
  \centering
  \includegraphics[width=0.80\linewidth]{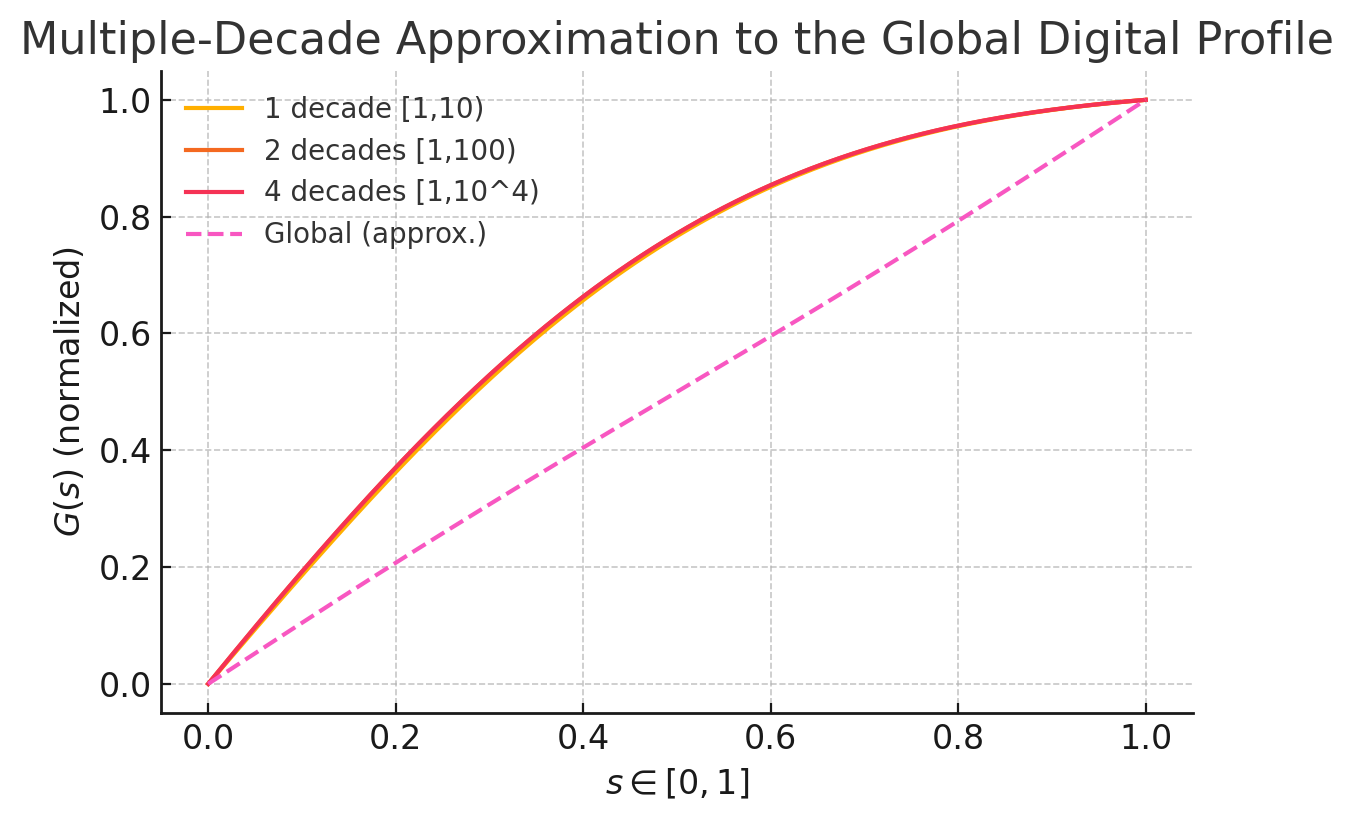}
  \caption{ Multiple-Decade Approximation to the Global Digital Profile.
  Curves show \(G_N(s)\) computed on \([1,10)\) (1 decade), \([1,100)\) (2 decades),
  and \([1,10^4)\) (4 decades), compared with an approximation to the global profile
  obtained by aggregating many decades and renormalizing. The windowed profiles
  converge rapidly toward the global digital signature as \(N\) increases. }
  \label{fig:multi_decade_G}
\end{figure}

Figure~\ref{fig:multi_decade_G} illustrates this effect for a lognormal model:
as \(N\) grows, \(G_N(s)\) stabilizes and closely tracks the global profile obtained
by summing over many decades and renormalizing. This provides a practical bridge
between finite samples on restricted magnitude ranges and the limiting behavior
over \((0,\infty)\).

\subsubsection*{6.1.2.3 Inverting the Windowed Profile: a Finite-Difference Equation}

Let \(m<n\) be integers (they may be negative) and consider the window of consecutive
decades \([10^{m},\,10^{n})\). For a CDF \(F\) on \((0,\infty)\) define
\begin{equation}
  \label{eq:windowG}
  G_{[m,n)}(s)
  :=
  \frac{\displaystyle\sum_{i=m}^{n-1}\!\bigl[F(10^{\,s+i})-F(10^{\,i})\bigr]}
       {\displaystyle F(10^{\,n})-F(10^{\,m})},
  \qquad s\in[0,1].
\end{equation}
This is the normalized cumulative digital profile of the distribution restricted to
\([10^{m},10^{n})\).

\paragraph{A convenient normalization and a finitedifference equation.}
Set
\[
\widetilde F(t):=\frac{F(10^{\,t})-F(10^{\,m})}{F(10^{\,n})-F(10^{\,m})},
\quad t\in\mathbb{R}.
\]
Then \eqref{eq:windowG} becomes
\[
G_{[m,n)}(s)
  = \sum_{i=m}^{n-1}\bigl[\widetilde F(s+i)-\widetilde F(i)\bigr].
\]
Let 
\(
c:=\frac{1}{L}\sum_{i=m}^{n-1}\widetilde F(i)
\)
with \(L:=n-m\), and define the centered function
\(V(t):=\widetilde F(t)-c\).
Noting that \(\sum_{i=m}^{n-1}V(i)=0\), we obtain the finite-difference (shift) equation
\begin{equation}
  \label{eq:diffeq}
  \boxed{\;
  \mathcal{S}_{m,n}V(s):=\sum_{i=m}^{n-1} V(s+i)\;=\;G_{[m,n)}(s),\qquad s\in[0,1].
  \;}
\end{equation}
Given \(G_{[m,n)}\), this is an inversion problem for \(V\); once \(V\) is found,
\(\widetilde F(t)=V(t)+c\), hence
\[
F(10^{\,t}) \;=\; F(10^{\,m}) + \bigl(F(10^{\,n})-F(10^{\,m})\bigr)\,\bigl(V(t)+c\bigr),
\quad t\in\mathbb{R}.
\]
Finally, recovering \(F(x)\) on the window amounts to evaluating the right-hand side at \(t=\log_{10}x\).

\paragraph{Structure of the solution space.}
The operator \(\mathcal{S}_{m,n}\) is the length-\(L\) discrete averaging (box-sum) over
integer shifts. Its kernel consists of functions \(W\) satisfying
\(\sum_{j=0}^{L-1}W(s+j)=0\).
A standard representation (via the characteristic equation
\(\sum_{j=0}^{L-1} r^{j}=0\)) shows that the homogeneous solutions are spanned by
\[
\cos\!\Bigl(\tfrac{2\pi q}{L}\,s\Bigr),\quad
\sin\!\Bigl(\tfrac{2\pi q}{L}\,s\Bigr), \qquad q=1,\dots,L-1,
\]
possibly with \(1\)-periodic amplitudes in \(\{s\}\) (the fractional part of \(s\)).
Thus the general solution of \eqref{eq:diffeq} can be written as
\begin{equation}
  \label{eq:generalV}
  V(s)\;=\;V_p(s) \;+\; \sum_{q=1}^{L-1}
  \Bigl(A_q(\{s\})\cos\tfrac{2\pi q}{L}s \;+\; B_q(\{s\})\sin\tfrac{2\pi q}{L}s\Bigr),
  \end{equation}
where \(V_p\) is any particular solution and \(A_q,B_q\) are arbitrary \(1\)-periodic functions.
Imposing monotonicity (CDF) or smoothness conditions typically forces \(A_q,B_q\) to vanish,
yielding a canonical (minimal-oscillation) solution.

\paragraph{Examples.}
\begin{enumerate}
  \item \emph{Single decade (\(L=1\)):} Here \(\mathcal{S}_{m,n}V(s)=V(s)=G_{[m,n)}(s)\).
        Thus \(V(s)=G_{[m,n)}(s)\) and
        \(\widetilde F(t)=c+G_{[m,n)}(t)\).
        In particular, if \(G_{[m,n)}(s)=s\) on \([1,10)\),
        then \(\widetilde F(10^{\,s})=s\), i.e.
        \(F(x)\propto \log_{10}x\) on \([1,10)\).

  \item \emph{Three decades (\(m=-1,\,n=2\Rightarrow L=3\)) with \(G_{[m,n)}(s)=s\):}
        Equation \eqref{eq:diffeq} takes the form
        \[
        V(s-1)+V(s)+V(s+1)=s.
        \]
        A particular solution is \(V_p(s)=s/3\) since
        \(\tfrac{1}{3}\bigl[(s-1)+s+(s+1)\bigr]=s\).
        The homogeneous equation has the characteristic roots
        \(e^{\pm 2\pi i/3}\), hence the full solution is
        \[
        V(s)=\frac{s}{3}
        \;+\;C_1(\{s\})\cos\!\Bigl(\frac{2\pi s}{3}\Bigr)
        \;+\;C_2(\{s\})\sin\!\Bigl(\frac{2\pi s}{3}\Bigr),
        \]
        with \(C_1,C_2\) arbitrary \(1\)-periodic functions.
        Choosing \(C_1\equiv C_2\equiv 0\) yields the simplest monotone reconstruction.
\end{enumerate}

\paragraph{Interpretation.}
The finite-difference relation in Equation~\eqref{eq:diffeq} reveals that the windowed profile \( G_{[m,n)}(s) \) is a discrete convolution of a shifted version of the function \( F \), mapped via base-10 logarithm, with a uniform (length-\(L\)) box kernel. This means that knowledge of the profile \( G_{[m,n)} \) over the interval \( s \in [0,1] \) determines the underlying distribution function \( F \) over the domain \([10^{m}, 10^{n})\), up to additive constants \( F(10^m) \), \( F(10^n) \), and any admissible homogeneous (oscillatory) components as described in Equation~\eqref{eq:generalV}. Imposing additional conditions on \( F \)—such as monotonicity and absolute continuity—serves to eliminate these oscillatory modes, yielding a unique, smooth, and non-oscillatory reconstruction of \( F \).

\subsection*{6.1.3 Illustration: $G(S)$ for the Normal Distribution}

As a concrete example, consider the normal (Gaussian) distribution:
\[
F(x) = \Phi\left(\frac{x - \mu}{\sigma}\right)
\]
where $\Phi$ is the standard normal CDF, $\mu$ is the mean, and $\sigma$ is the standard deviation. The figures below show $G(S)$ for several values of $\mu$, with $\sigma = 1$, $A = 1$, $B = 10$.

\begin{figure}[H]
    \centering
    \begin{subfigure}{0.45\textwidth}
        \includegraphics[width=\linewidth]{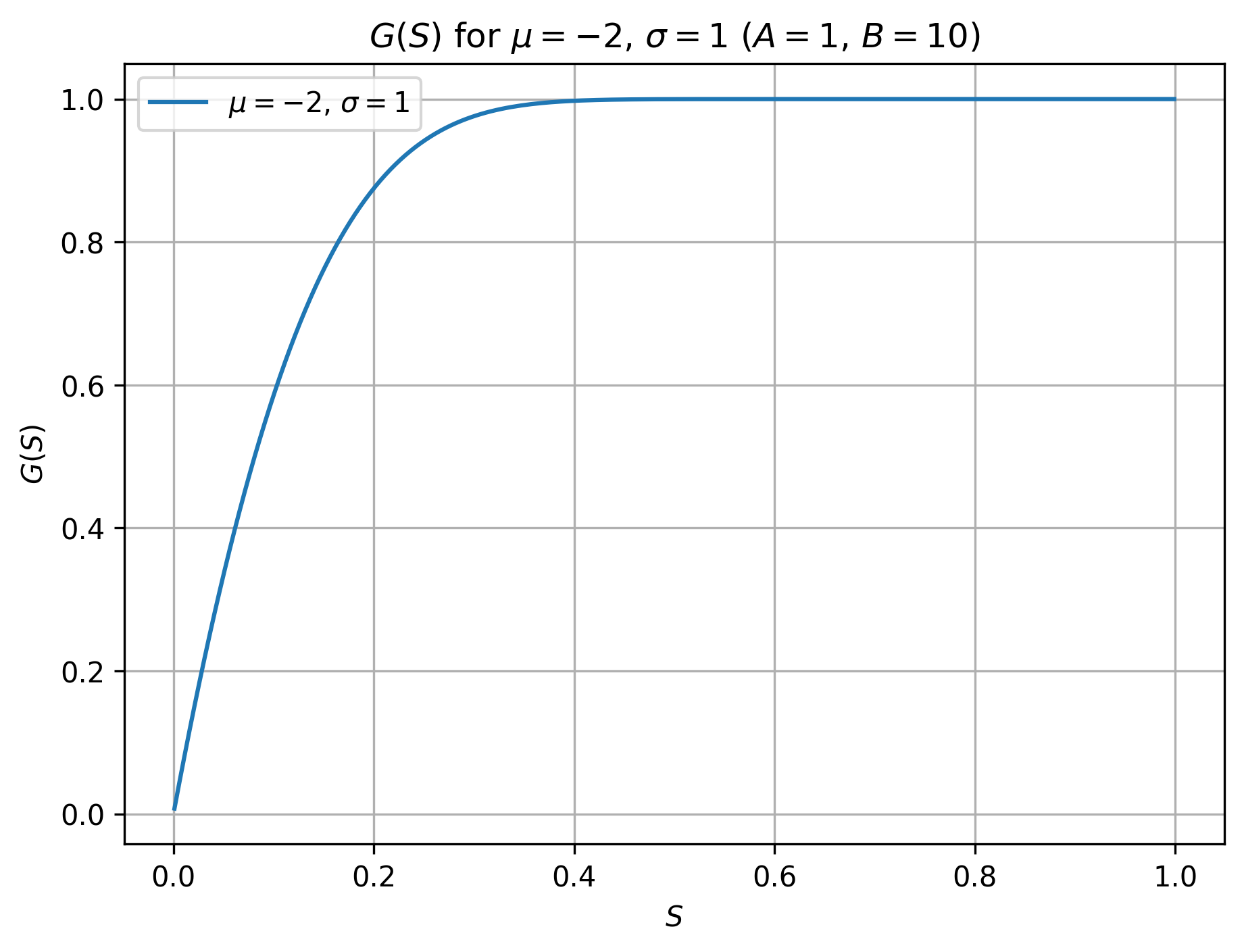}
        \caption{$\mu = -2$, $\sigma = 1$}
    \end{subfigure}
    \hfill
    \begin{subfigure}{0.45\textwidth}
        \includegraphics[width=\linewidth]{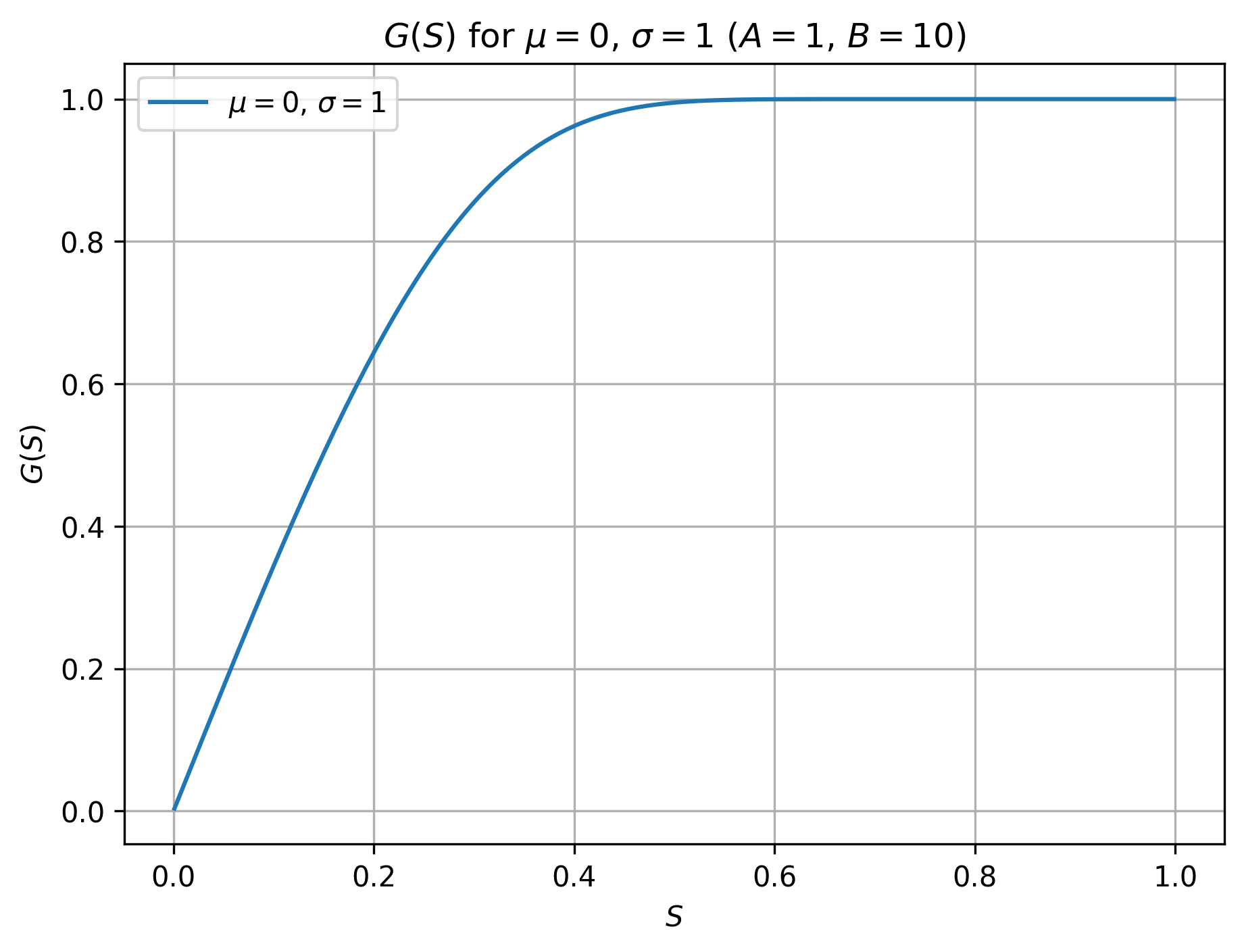}
        \caption{$\mu = 0$, $\sigma = 1$}
    \end{subfigure}
    \vspace{0.5cm}

    \begin{subfigure}{0.45\textwidth}
        \includegraphics[width=\linewidth]{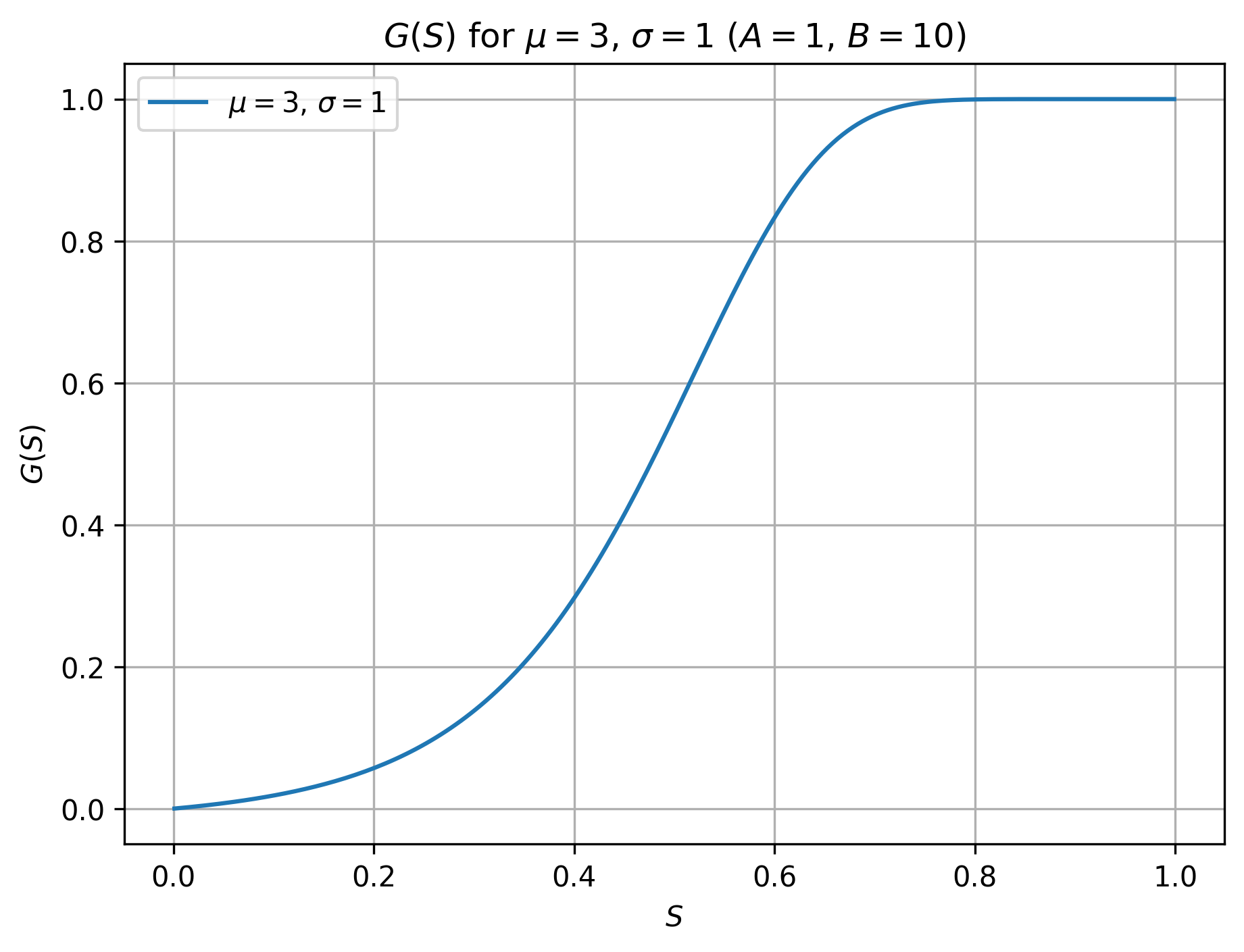}
        \caption{$\mu = 3$, $\sigma = 1$}
    \end{subfigure}
    \hfill
    \begin{subfigure}{0.45\textwidth}
        \includegraphics[width=\linewidth]{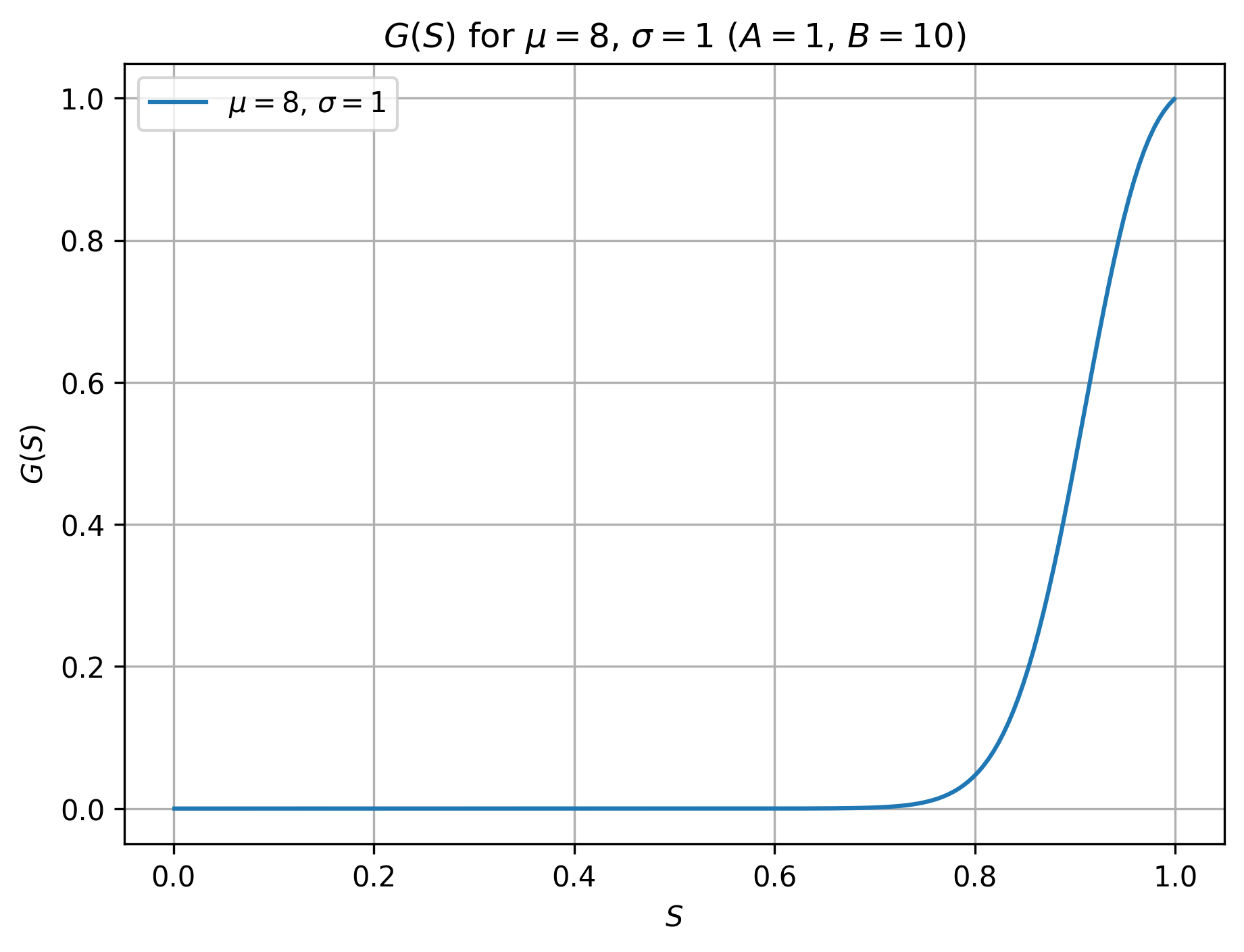}
        \caption{$\mu = 8$, $\sigma = 1$}
    \end{subfigure}
    \vspace{0.5cm}

    \begin{subfigure}{0.45\textwidth}
        \includegraphics[width=\linewidth]{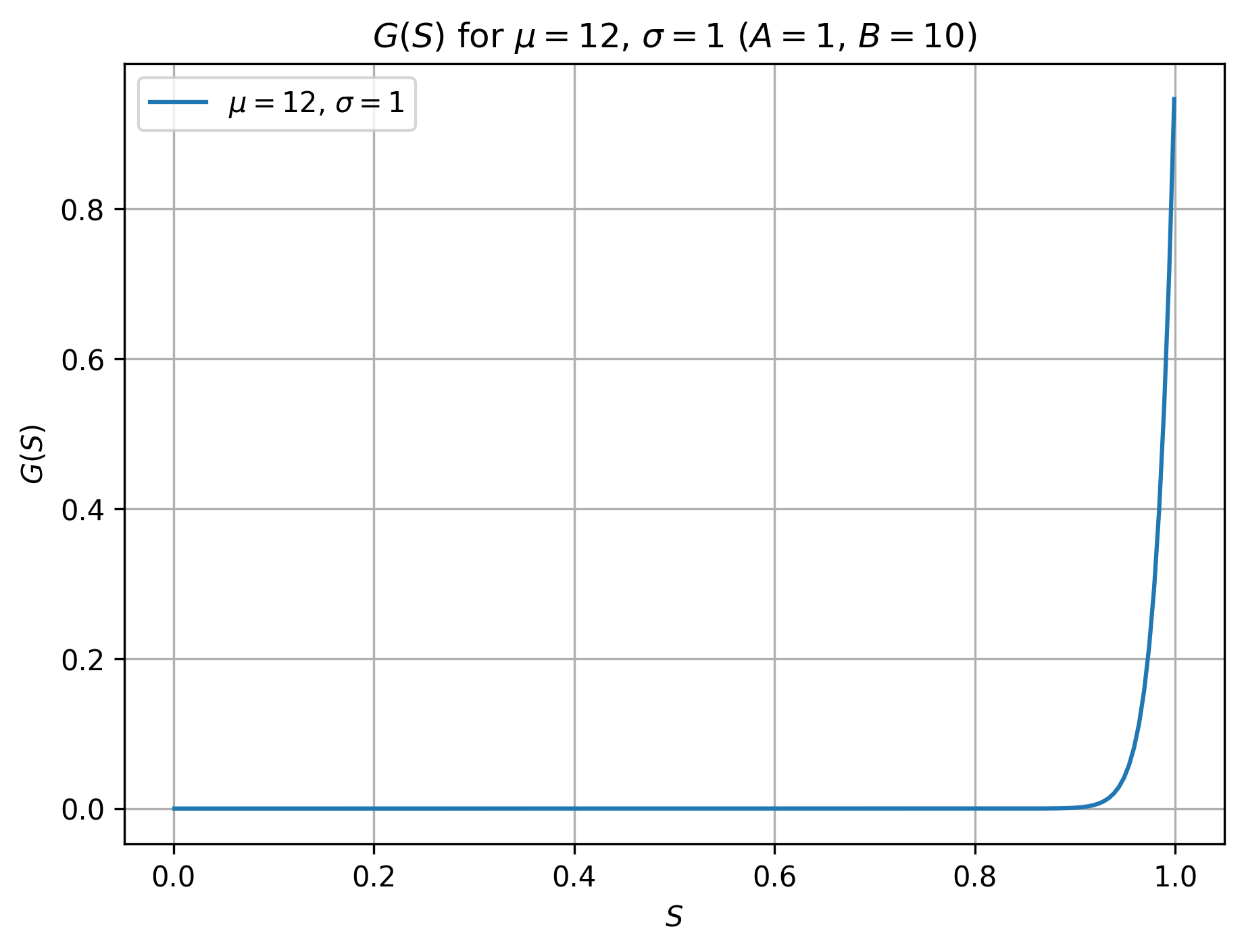}
        \caption{$\mu = 12$, $\sigma = 1$}
    \end{subfigure}
    \caption{Plots of $G(S)$ for different values of $\mu$ for the normal distribution ($A=1$, $B=10$, $\sigma=1$).}
    \label{fig:gs_normal}
\end{figure}

\begin{figure}[H]
    \centering
    \includegraphics[width=0.7\textwidth]{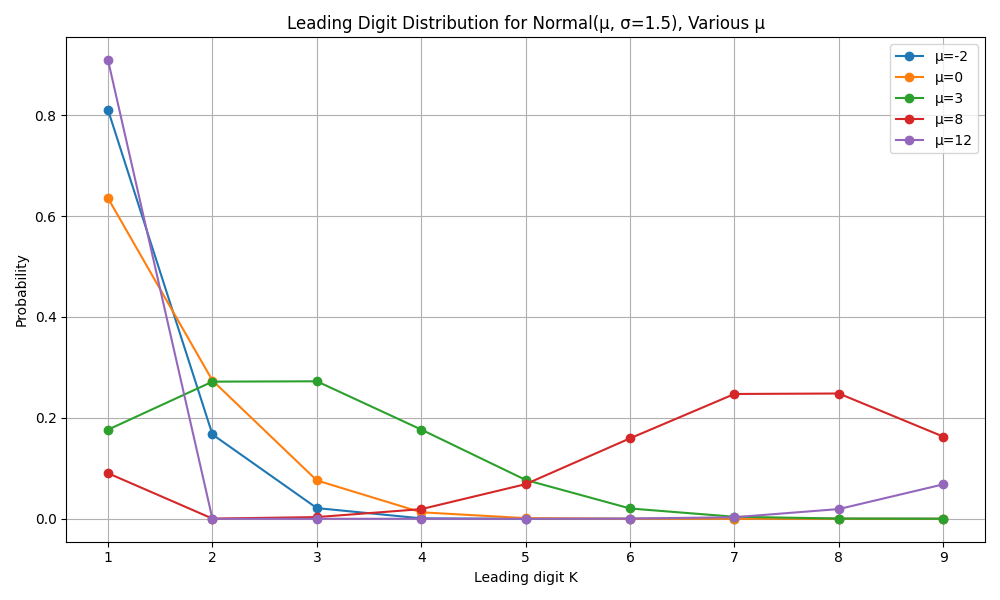}
    \caption{Leading digit profile for samples from the normal distribution $N(\mu, \sigma^2)$ with different $\mu$, $\sigma = 1.5$.}
    \label{fig:ldpf_normal_mu}
\end{figure}

\vspace{0.3cm}
\noindent
\textit{Remark.} The framework above will be extended to general intervals and compared with theoretical models, including Benford's Law, in subsequent subsections.

\subsubsection*{6.1.4 \quad The Role of Interval Endpoints: Complete and Incomplete Leading Digit Sets}

An important aspect of leading digit analysis is the choice of the interval $[A, B]$ from which numbers are sampled. Specifically, the coverage of all possible leading digits within the interval is directly affected by the values of $A$ and $B$.

Consider first the case where $A = 1$ and $B = 10$. This interval is precisely one decade in the decimal system, containing all numbers of the form $d \times 10^{s}$ with $d \in \{1, \ldots, 9\}$ and $0 \leq s < 1$. Consequently, every possible leading digit $d$ occurs somewhere in $[1, 10)$, guaranteeing a complete set of first digits.

If, however, the upper endpoint $B$ is not a power of ten—say, $B = \pi$ or another arbitrary (possibly rational or irrational) value—then the situation changes fundamentally. For example, in $[1, \pi)$, only numbers with leading digit $1$, $2$, or $3$ can be present, since $\pi \approx 3.14$. Digits $4$ through $9$ are necessarily absent from any sample restricted to this interval. Similarly, for an interval $[1, b)$ where $b$ is not a decade endpoint (e.g., $b=7$), only numbers with leading digits $1$ through $6$ are possible.

This incomplete coverage of possible digits arises from the structure of the decimal system: only intervals of the form $[1, 10^k)$, for integer $k$, contain all leading digits. For all other choices of $B$, the set of accessible leading digits is restricted, and may even depend in subtle ways on the decimal expansion of $B$. In particular, for irrational endpoints, the issue of coverage becomes nontrivial and may have theoretical and practical consequences.

To further illustrate this point, we can consider the interval $[e, \pi)$, where both endpoints are irrational. In this case, only numbers whose leading digit falls within the range covered by $e \approx 2.718$ and $\pi \approx 3.1416$ will appear, meaning only certain digits (such as $2$ and $3$) may be present as the leading digit, while others (such as $1$, $4$, $5$, ..., $9$) will be absent.

\begin{figure}[H]
    \centering
    \includegraphics[width=0.6\textwidth]{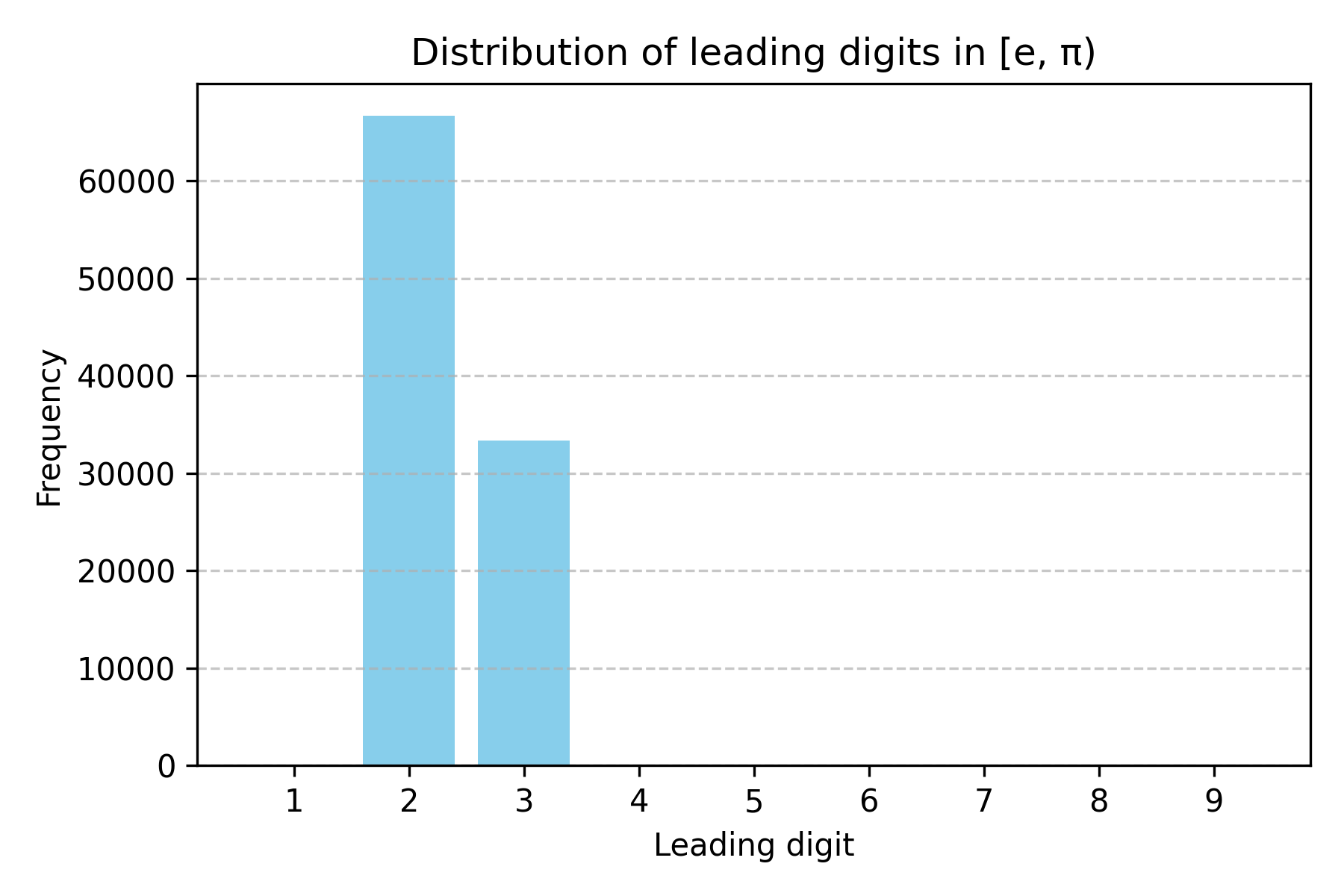}
    \caption{Histogram of leading digits for 100{,}000 numbers sampled uniformly from the interval $[e, \pi)$. Only digits 2 and 3 appear as the leading digit in this range.}
    \label{fig:leading_digit_e_pi}
\end{figure}

Understanding these constraints is essential when designing experiments or interpreting data on leading digit distributions. It highlights why, in foundational studies and in Benford’s Law itself, the use of full decades ($[1, 10)$, $[10, 100)$, etc.) is standard.

\vspace{0.2cm}
\noindent
\textit{In the following sections, we will demonstrate these effects explicitly by comparing distributions of leading digits in intervals with both decade and non-decade endpoints.}

\subsection*{6.2 Ratio of Uniforms and Digital Signature}

A \textbf{full-support density} on the positive real axis is a probability density function (PDF) $f(z)$ that is strictly positive for all $z > 0$. Such distributions are central in the analysis of digital (leading digit) phenomena, because they allow the random variable to assume any positive value, thus covering all possible orders of magnitude.

A fundamental example is the distribution of the ratio $z = y/x$, where $x$ and $y$ are independent and uniformly distributed on $(0,1)$. The resulting PDF, well known from classical probability theory , is
\[
f(z) = 
\begin{cases}
\frac{1}{2}, & 0 < z < 1 \\[1em]
\frac{1}{2z^{2}}, & z \geq 1
\end{cases}
\]
This density assigns nonzero probability to every positive $z$, featuring a constant part for $z < 1$ and a slowly decaying, scale-invariant tail for $z \geq 1$.

\textbf{Our goal} is to analyze the digital signature (leading digit profile) of such distributions using the jet function formalism. Specifically, we compute the following integral:
\[
G(S) = \int_0^{\infty} f(z) \, M(S, z) \, dz,
\]
where $M(S, z) = -\lfloor \{\log_{10} z\} - S\rfloor$  is the logarithmic folding function, and $\{\cdot\}$ denotes the fractional part.

\vspace{0.2cm}
\noindent
\textit{By direct computation, we find that:}
\[
G(S) = \frac{1}{2} + \frac{10^{S}}{18} - \frac{5 \cdot 10^{-S}}{9}
\]
This explicit analytic formula describes the leading digit profile for the ratio of two independent uniform random variables on $(0,1)$. It provides a complete digital signature over the entire positive real line.

\vspace{0.2cm}
\noindent
\textit{Remark:} The key difference from previous cases is that here the integration is performed over the full interval $(0, \infty)$, without artificial truncation. This reveals the true digital nature of the underlying full-support density.

\vspace{0.3cm}
\noindent\textbf{Explicit formula for the leading digit probability.}

The probability $\rho(k, n)$ that the leading digit of $z = y/x$ (with $x, y$ independent and uniformly distributed on $(0,1)$) equals $k$ at scale $n$ is given by:
\[
\rho(k, n) = \frac{10^{-n}}{18} - \frac{5}{9} \cdot 10^n \left( \frac{1}{k+1} - \frac{1}{k} \right)
\]
where $n = \lfloor \log_{10} k \rfloor$.

This formula allows direct computation of the probability for any leading digit $k$ and scale $n$ in the full-support ratio distribution.

\subsection*{6.3 Alternative Underlying Distributions for the Same LDPF}

It is important to note that the same leading digit profile function (LDPF), here denoted as $G_{\text{RTS}}(s)$, can arise from very different underlying probability distributions. For example, the LDPF for the ratio of two independent Uniform$(0,1)$ random variables, which we have previously found in analytic form, may also correspond to an entirely different distribution defined on the interval $[1, 10]$.

Suppose we seek a density $f(x)$ supported on $[1,10]$ such that
\[
G(s) = \frac{F(10^s) - F(1)}{F(10) - F(1)}
\]
matches a given target function, e.g., the analytic LDPF obtained for the ratio of uniforms:
\[
G(s) = \frac{1}{18} \cdot 10^s - \frac{5}{9} \cdot 10^{-s}+\frac{1}{2} , \qquad s \in [0,1].
\]

To reconstruct the corresponding density, recall that $F(x)$ is the cumulative distribution function, and $f(x) = \frac{dF}{dx}$. Expressing $G(s)$ in terms of $x=10^s$:
\[
G(s) = \frac{F(x) - F(1)}{F(10) - F(1)}
\implies F(x) = (F(10) - F(1)) G(s) + F(1), \qquad x \in [1,10].
\]
Taking the derivative with respect to $x$:
\[
f(x) = \frac{dF}{dx} = (F(10) - F(1)) \frac{dG}{dx}
\]
where $G(s)$ is now viewed as a function of $x$ via $s = \log_{10} x$.

Since $s = \log_{10} x$, $ds/dx = \frac{1}{x \ln 10}$, and thus:
\[
\frac{dG}{dx} = \frac{dG}{ds} \cdot \frac{ds}{dx} = G'(s) \cdot \frac{1}{x \ln 10}
\]

Therefore,
\[
f(x) = (F(10) - F(1)) \frac{1}{x \ln 10} G'(s), \qquad s = \log_{10} x.
\]

By plugging in the analytic expression for $G(s)$ and differentiating, we obtain an explicit form for $f(x)$ on $[1,10]$ which produces exactly the same LDPF as the ratio of uniforms.

Therefore, the corresponding PDF $f(x)$ on $[1,10]$ that produces exactly the same leading digit profile (LDPF) is
\[
f(x) = \frac{1}{18} + \frac{5}{9x^2}, \qquad x \in [1,10].
\]

with $(F(10) - F(1))=1$.
This function is elementary and does not require any logarithmic differentiation or special normalization constants.

\vspace{0.2cm}
\noindent
\textit{Remark:} This demonstrates that LDPF does not uniquely determine the underlying distribution; different PDFs, possibly defined on entirely different intervals, can share the same leading digit profile function.

\subsection*{6.4. Leading Digit Profiles for Products of Independent Uniform Variables}

For the product $Z = X_1 X_2$ of two independent random variables $X_1, X_2$ uniformly distributed on $(0,1)$, the cumulative distribution function (CDF) is
\[
F_Z(z) = z \left(1 - \ln z \right), \qquad z \in (0,1)
\]
The corresponding probability density function (PDF) is
\[
f_Z(z) = -\ln z, \qquad z \in (0,1)
\]
This explicit form for the CDF and PDF allows us to compute digital profiles and analyze leading digit statistics for products of independent random variables.

We will use these explicit forms to compute the leading digit profile function (LDPF) and to further analyze the digital properties of products of independent random variables.

\noindent
\textbf{Example 1: Analytic LDPF as an Integral}

The leading digit profile function (LDPF) $G(s)$ for the product $Z = X_1 X_2$ of two independent Uniform$(0,1)$ variables can be expressed as the following integral:
\[
G(s) = \int_0^1 \left[ -\ln z \right] \, M(s, z) \, dz,
\]
where $M(s, z)$ is the logarithmic folding function (or indicator), and $s \in [0,1]$.

Evaluating this integral over the interval $z \in (0,1)$, and performing normalization and scaling, yields the analytic expression
\[
G(s) = 10^s (A s + B) - B,
\]
with constants $A$ and $B$ determined by the normalization conditions $G(0) = 0$, $G(1) = 1$.

\[
A = -\frac{\ln 10}{9}, \qquad B = \frac{1}{9} + \frac{10 \ln 10}{81}.
\]

This formula ensures correct normalization and scaling of the digital profile over one decade.

This demonstrates the direct connection between the probabilistic integral and the explicit analytic formula for the digital profile.

\noindent
\textbf{Example 2: CDF-based construction on $[0.1, 1]$}

Suppose $Z = X_1 X_2$, where $X_1, X_2$ are independent and uniformly distributed on $(0,1)$. The cumulative distribution function is
\[
F_Z(z) = z \left(1 - \ln z \right), \qquad z \in (0,1).
\]

For leading digit analysis on the interval $[0.1, 1]$, we define the profile function as
\[
G(s) = \frac{F_Z(10^{-1 + s}) - F_Z(0.1)}{F_Z(1) - F_Z(0.1)},
\]
for $s \in [0,1]$.

Explicitly:
\[
F_Z(10^{-1 + s}) = 10^{-1 + s} \left[ 1 - \ln (10^{-1 + s}) \right] = 10^{-1 + s} \left[ 1 + (1 - s) \ln 10 \right].
\]
Similarly,
\[
F_Z(0.1) = 0.1 \left(1 - \ln 0.1\right) = 0.1 (1 + \ln 10),
\]
\[
F_Z(1) = 1 \cdot (1 - \ln 1) = 1.
\]

Thus, the leading digit profile function becomes:
\[
G(s) = \frac{10^{s} \left[1 + (1 - s)\ln 10\right] - (1 + \ln 10)}{9 -  \ln 10}
\]
for $s \in [0, 1]$.

\textit{This explicit formula allows direct computation of the digital profile for the product of two uniforms on $[0.1, 1]$.}

\vspace{0.3cm}
\noindent
%
%
%

\begin{remark}[Limiting case: product of uniforms and Benford’s law]
Let $U_1,\dots,U_N \sim \mathrm{Unif}(0,1)$ be independent and set
\[
Z_N=\prod_{i=1}^N U_i .
\]
Then
\[
Y_N:=-\ln Z_N=\sum_{i=1}^N(-\ln U_i)\sim\mathrm{Gamma}(N,1),
\]
so that the density of $Z_N$ on $(0,1)$ is
\[
f_{Z_N}(z)=\frac{(-\ln z)^{N-1}}{\Gamma(N)}=\frac{(-\ln z)^{N-1}}{(N-1)!},
\qquad 0<z<1 .
\]
Moreover,
\[
\log_{10} Z_N=\frac{1}{\ln 10}\sum_{i=1}^N \ln U_i .
\]
Because $\ln U_i$ has a continuous, nonlattice distribution, the fractional
parts $\{\log_{10} Z_N\}$ converge in distribution to $\mathrm{Unif}[0,1)$ as
$N\to\infty$ (standard Fourier/Weyl equidistribution argument). Consequently the
leading-digit/windowed profile
\[
G_N(s)=\Pr\bigl(\{\log_{10} Z_N\}\le s\bigr)\;\longrightarrow\; s,
\qquad s\in[0,1),
\]
i.e., Benford’s law is recovered in the product limit.
\end{remark}

\subsection*{6.5 Power Law Distribution: Analytical Expression and Special Cases}

Let
\[
f(y) = \frac{m\, y^{m-1}}{b^m}, \quad y \in [0, b],\quad m > 0
\]
with
\[
G(s) = \int_0^b \frac{m\, x^{m-1}}{b^m} M(s, x)\, dx, \qquad M(s,x) = -\lfloor\, \{ \log_{10} x \} - s \,\rfloor
\]
where \( \{x\} = x - \lfloor x \rfloor \).

\subsubsection*{6.5.1 The general case}

Using the abbreviations:
\newcommand{\logten}[1]{\log_{10}\!\left(#1\right)}
\newcommand{\fract}[1]{\left\{\,#1\,\right\}}       
\newcommand{\floor}[1]{\left\lfloor #1\right\rfloor}

\[
\begin{aligned}
q      &= \logten{b},         &\qquad q_0 &= \fract{q},          &\qquad q_1 &= \floor{q}, \\[2pt]
r      &= -a + q,             &        r_0 &= \fract{r},          &        r_1 &= \floor{r}.
\end{aligned}
\]

With these abbreviations the general windowed profile for \(m\ge -1\) becomes
\[
G(a,b,m)\;=\;
-\,r_1 \;+\; q_1
\;-\;
\frac{10^{m}\,10^{-m q_0}}{10^{m}-1}
\;+\;
\frac{10^{m}\,10^{-m r_0}}{10^{m}-1}.
\]

When $b = 10^n$, then $\log_{10} b$ is integer and thus $\{\log_{10} b\} = 0$. The result simplifies to (write the explicit form or simply say: the leading digit profile is strictly uniform due to exact power-of-ten cutoff).

Then
\[
G(s\,;\,b=10^n,m)
=\frac{\displaystyle \sum_{i=0}^{m-1}\!\bigl[F(10^{s+i})-F(10^{i})\bigr]}
        {F(10^{m})-F(1)}
=\boxed{\;\frac{10^{m s}-1}{\,10^{m}-1}\;},\qquad s\in[0,1).
\]

\subsubsection*{6.5.2 The case $m=1$ and $b \neq 10^n$}

For $m=1$, the density is uniform, but $b$ is not an integer power of 10.
The leading digit profile is:
\[
G(s) = \frac{10 \cdot 10^{-U_1}}{9} - \frac{10 \cdot 10^{-V_1}}{9} - U
\]
where
\begin{align*}
    U   &= \left\lfloor-s + \{ \log_{10} b \}  \right\rfloor, \\
    U_1 &= \{ -s + \{ \log_{10} b \} \}, \\
    V_1 &= \{ \log_{10} b \}
\end{align*}
and $\{x\} = x - \lfloor x \rfloor$.

\medskip
Below are figures of $G(s)$ for several choices of $\{\log_{10} b\} = 0.1, 0.3, 0.6, 0.8$.

\begin{figure}[H]
    \centering
    \includegraphics[width=0.8\textwidth]{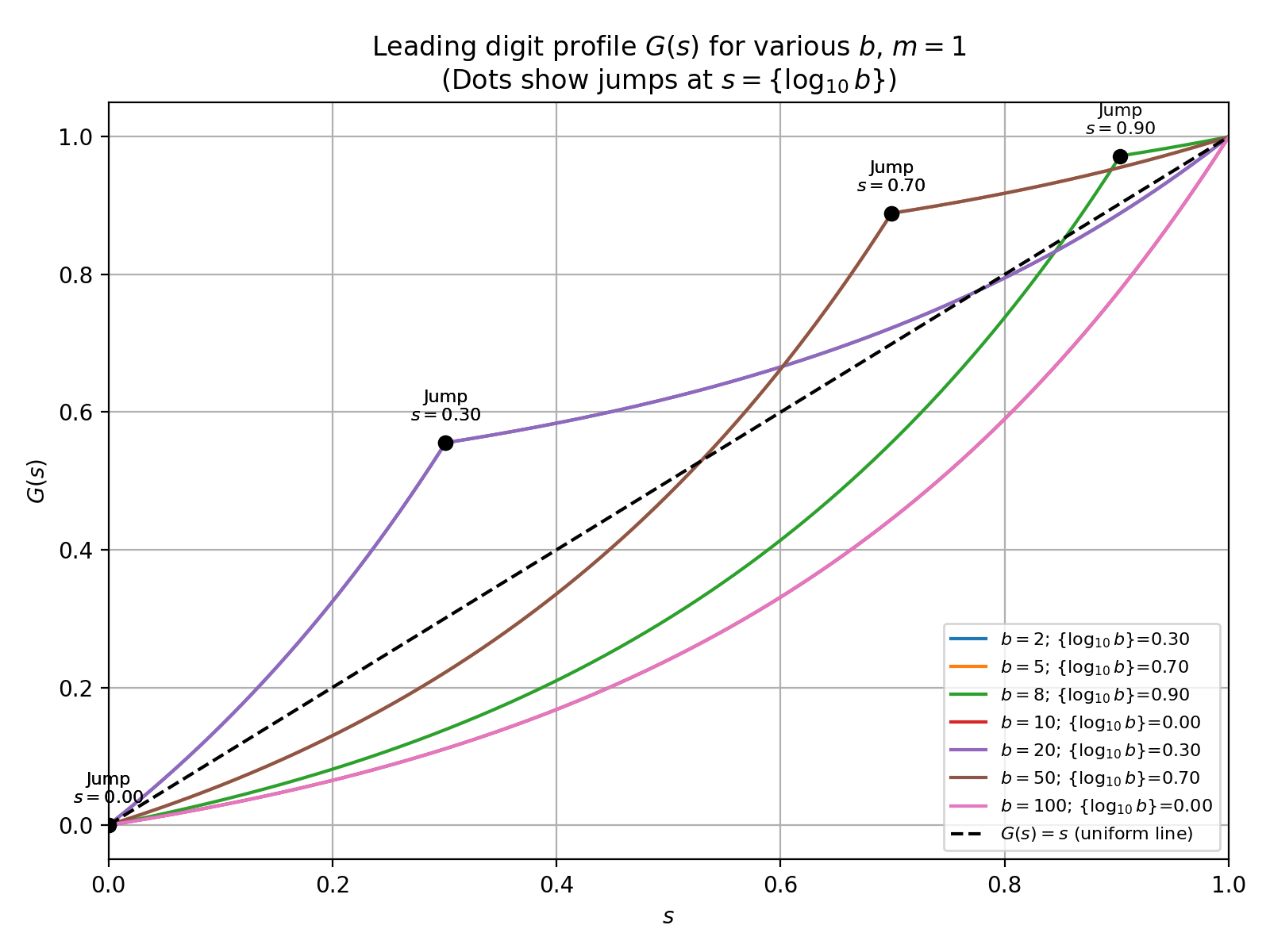}
    \caption{
        Leading digit profile function \(G(s)\) for the power law with \(m=1\) and several upper bounds \(b\).
        Each curve corresponds to a different value of \(b\), with the corresponding fractional part \(\{\log_{10} b\}\) indicated in the legend.
        The black dots and "Jump" labels mark the points \(s = \{\log_{10} b\}\), where \(G(s)\) exhibits a jump discontinuity, corresponding to the cut-off of the support at \(x=b\).
        For \(b=10^k\) (\(\{\log_{10} b\}=0\)), \(G(s)\) coincides with the uniform distribution \(G(s)=s\) (shown as a dashed line).
        As \(b\) varies between powers of ten, the position and magnitude of the jump shift, illustrating how the profile depends on the upper bound of the distribution.
    }
    \label{fig:GG_all_b_on_one_plot}
\end{figure}

\textbf{Explanation.}
Figure~\ref{fig:GG_all_b_on_one_plot} demonstrates the leading digit profile function \(G(s)\) for samples drawn from a power law with exponent \(m=1\) over the interval \([0, b]\), for several values of \(b\).
Each colored curve represents a different \(b\), and the corresponding fractional part \(\{\log_{10} b\}\) is shown in the legend.
A distinctive feature of these curves is the presence of a jump at \(s = \{\log_{10} b\}\), which is clearly marked on the plot.
This jump arises because at that specific value of \(s\), the support of the distribution is abruptly truncated by the upper bound \(b\).
The magnitude and location of the jump depend on \(b\), and as \(b\) approaches an exact power of ten (\(b = 10^k\)), the jump moves to \(s = 0\) and disappears, so the profile becomes perfectly linear (\(G(s) = s\)), as shown by the dashed line.
This illustrates the sensitivity of the leading digit profile to the upper bound, and highlights the unique case of the uniform profile for powers of ten.

\textbf{Note.}  
For the special case \(m=0\), the result is always \(G(s) = s\), corresponding to the uniform (Benford-like) case. However, there are many distributions whose leading digit profile is not of Benford form. The degree of deviation from Benford’s Law is precisely reflected in how much \(G(s)\) differs from the identity line \(s\). The jumps at \(s = \{\log_{10} b\}\) are visual markers of this deviation for different choices of \(b\).

\subsubsection*{6.5.3 Numerical Results and Periodicity of $\rho(k, b, m)$}

This subsection presents a numerical investigation of the leading digit frequency function $\rho(k, b, m)$ for the power law distribution on $[1, b]$, with several values of the exponent $m$ and leading digit $k$.

The function is defined as
\[
\rho(k, b, m) = \lfloor \log_{10}(k+1) \rfloor - \lfloor \log_{10}(k) \rfloor + G(\{\log_{10}(k+1)\}, b, m) - G(\{\log_{10}(k)\}, b, m)
\]
where $G(s, b, m)$ is the leading digit profile function described earlier, and $\{x\} = x - \lfloor x \rfloor$ is the fractional part.

The following figures illustrate the dependence of $\rho(k, b, m)$ on the upper bound $b$ and exponent $m$ for three representative digits $k=1,3,8$. Each plot shows how the frequency of the leading digit changes as $b$ varies from $1$ to $10$ for a fixed $m$.

\begin{figure}[H]
    \centering
    \includegraphics[width=0.8\textwidth]{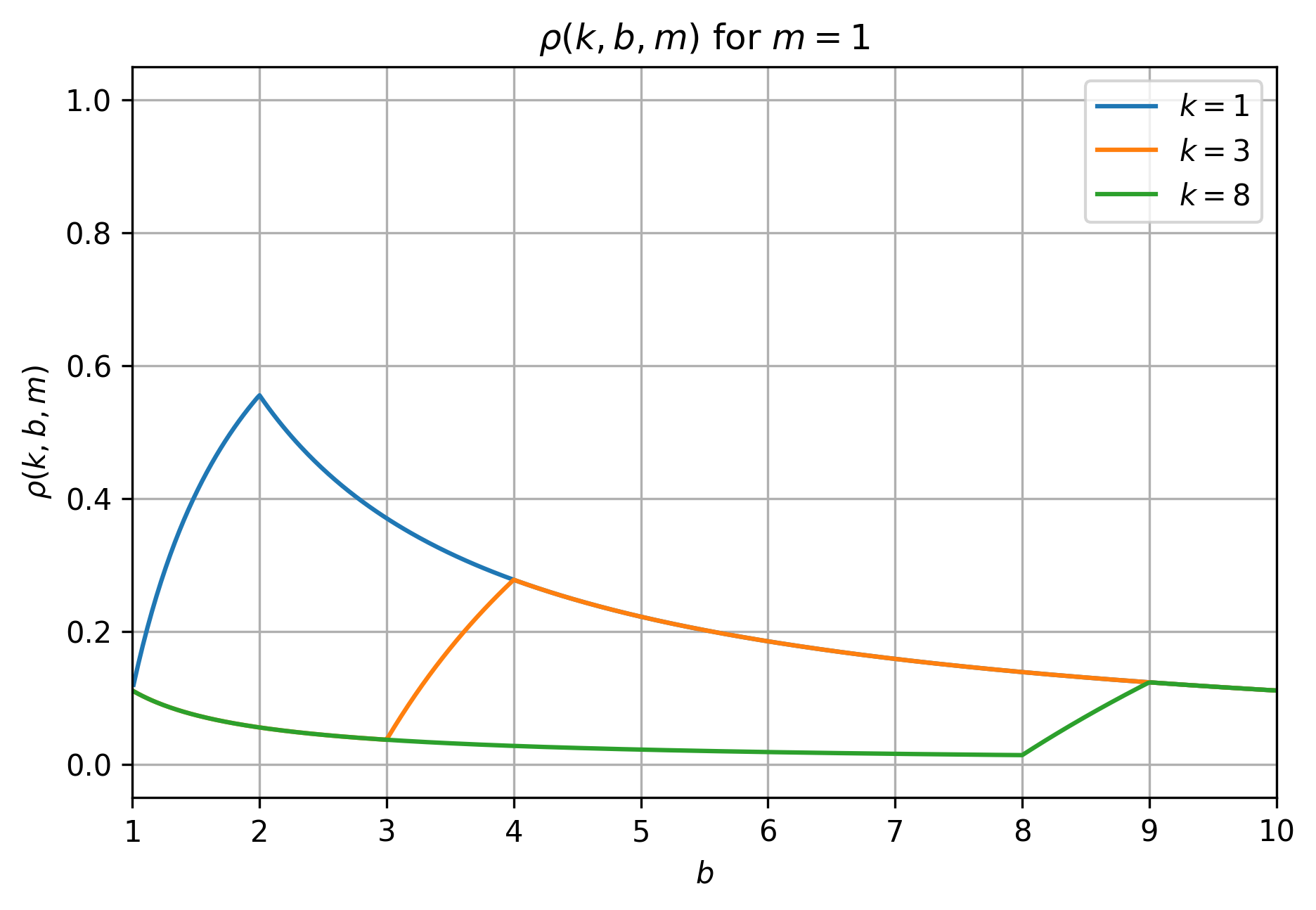}
    \caption{
        The function $\rho(k, b, m)$ for $m=1$ and $k=1,3,8$, as $b$ runs from $1$ to $10$. This case corresponds to the uniform distribution on $[1, b]$. For $b=10$, the digit frequencies approach Benford's Law. Between powers of ten, the profiles exhibit a characteristic “wavy” structure.
    }
    \label{fig:rho_k_1_3_8_m_1}
\end{figure}

\begin{figure}[H]
    \centering
    \includegraphics[width=0.8\textwidth]{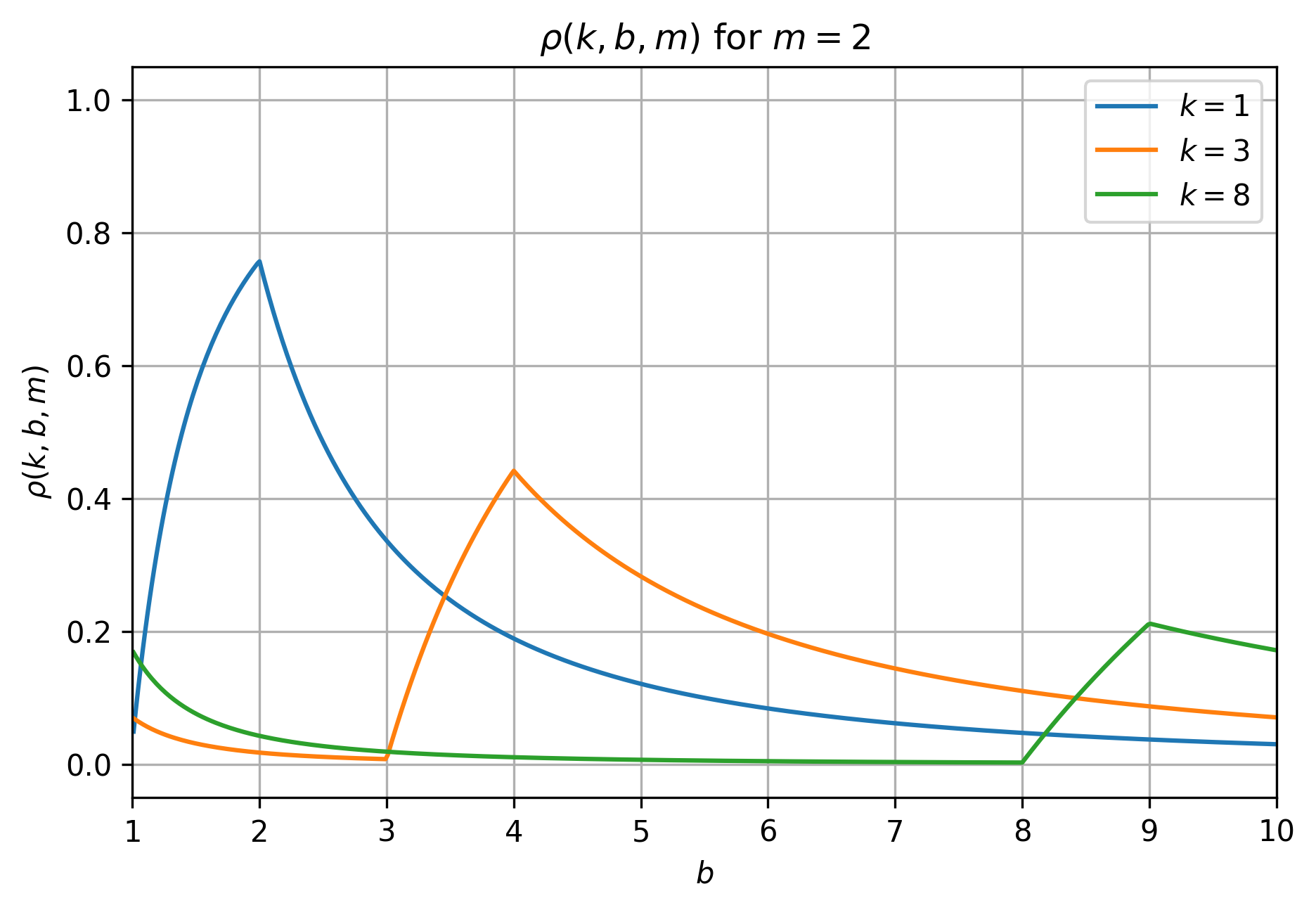}
    \caption{
        The function $\rho(k, b, m)$ for $m=2$ and $k=1,3,8$. Here, the digit frequencies become more concentrated due to the stronger bias toward larger values in the distribution. The “wavy” pattern persists and demonstrates the effect of the upper bound $b$.
    }
    \label{fig:rho_k_1_3_8_m_2}
\end{figure}

\begin{figure}[H]
    \centering
    \includegraphics[width=0.8\textwidth]{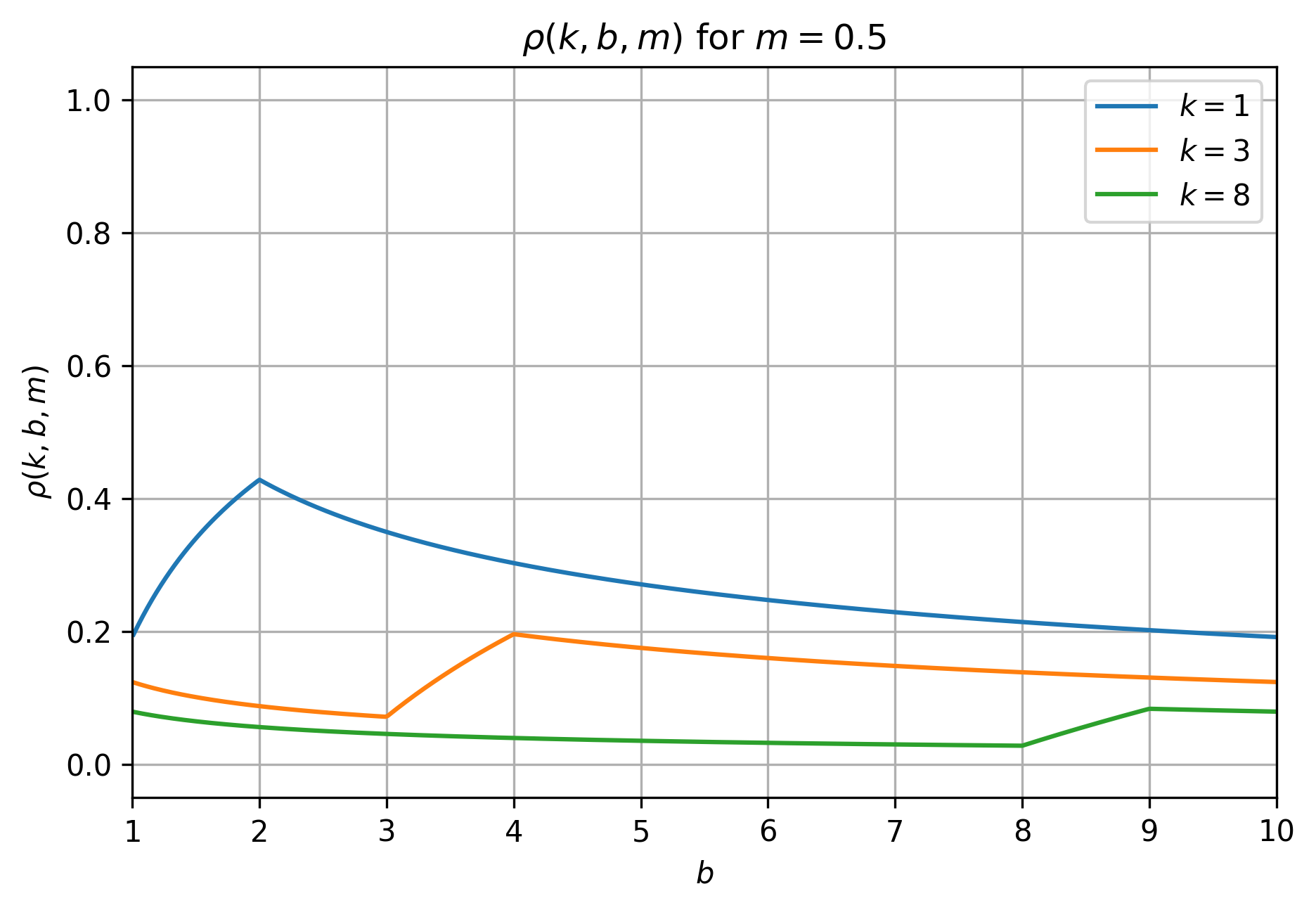}
    \caption{
        The function $\rho(k, b, m)$ for $m=0.5$ and $k=1,3,8$. With $m < 1$, the distribution is skewed toward lower values, which is reflected in the digit frequencies. The periodic, oscillatory nature of the profiles remains visible.
    }
    \label{fig:rho_k_1_3_8_m_0_5}
\end{figure}

\textbf{Discussion and Periodicity.}

A striking feature of these results is the \emph{periodic} behavior of $\rho(k, b, m)$ as a function of $\log_{10} b$. Specifically, as the upper bound $b$ increases by a factor of $10$ (that is, as $\log_{10} b$ increases by $1$), the shape of each curve is exactly repeated. This periodicity originates from the logarithmic structure of the digit distribution and means that, in each new decade, the profiles for all $k$ are reproduced. The “waves” or oscillations seen in the figures illustrate how the leading digit statistics cycle through all possible forms as $b$ traverses from one order of magnitude to the next.

For $m=1$, the case of the uniform distribution, the digit frequencies approach Benford’s Law exactly as $b \to 10$, and repeat the characteristic deviation for each new decade. For $m>1$, the distribution is more heavily weighted toward large values, and for $m<1$ toward small values; but in all cases, the periodic pattern remains—a fundamental and visually striking property of the digital statistics for power-law type distributions.

In addition, the calculated function exhibits clearly defined periodic behavior with respect to the decimal logarithm of $b$ with a period of 1. This means that the values of the function's \textbf{Maximum} and \textbf{Minimum} are determined by the position of the fractional part of the decimal logarithm of $k$.

The \textbf{Maximum} is achieved at the point where the fractional part of $\log_{10}( k+1)$  coincides with the fractional part of $\log_{10} b$ in the interval $[0, 1]$:
\[
\{\log_{10} (k + 1)\} = \{\log_{10} b\}
\]

The \textbf{Minimum} is achieved when the fractional part of $\log_{10} k$ coincides with the fractional part of $\log_{10} b$:
\[
\{\log_{10} (k )\} = \{\log_{10} b\}.
\]

The maximum and minimum values of the function can be explicitly written as:
\[
\rho(k)_{\mathrm{max}} = \rho\left(k,\, b = 10^{\left\{ \frac{\ln(k+1)}{\ln(10)} \right\}},\, m \right)
= \frac{ - (k+1)^{-m} k^m + 1 }{ 10^m - 1 } \cdot 10^m
\]
\[
\rho(k)_{\mathrm{min}} = \rho\left(k,\, b = 10^{\left\{ \frac{\ln k}{\ln(10)} \right\}},\, m \right)
= \frac{ (k+1)^m k^{-m} - 1 }{ 10^m - 1 }
\]

For the special case \( m = 1 \), these reduce to simple expressions:
\[
\rho(k)_{\mathrm{max}} = \frac{10}{9} \cdot \frac{1}{k+1}
\qquad
\rho(k)_{\mathrm{min}} = \frac{10}{9} \cdot \frac{1}{k}
\]

\textbf{Extension to Finite Intervals.}
The analysis above naturally leads to a practical question: what if, instead of considering the entire positive axis, we wish to study the leading digit statistics for a power-law distribution on a finite interval $[a, b]$?

This can be done directly and elegantly using a normalization formula. Suppose we already know the cumulative digital profile $G(s, [0, b], m)$ for the interval $[0, b]$, where $m$ is the power-law exponent. Then, the corresponding profile for any finite subinterval $[a, b]$ is given by
\begin{equation}
G(s, [a, b], m) = \frac{b^m\, G(s, [0, b], m) - a^m\, G(s, [0, a], m)}{b^m - a^m}
\end{equation}
Here, $G(s, [a, b], m)$ is the cumulative digital profile on $[a, b]$, and $G(s, [0, b], m)$, $G(s, [0, a], m)$ are the corresponding profiles computed for the intervals $[0, b]$ and $[0, a]$.

\textit{Intuitive explanation.}
This formula is a direct consequence of the properties of power-law distributions. Since the density is proportional to $x^{m-1}$, the contributions from $[0, b]$ and $[0, a]$ are scaled by $b^m$ and $a^m$, respectively. Subtracting the contribution up to $a$ and properly normalizing by the interval size, we obtain the leading digit statistics for any finite interval. This is especially useful for empirical data, where observations are always bounded.

With this extension, we are able to generalize the leading digit analysis to any desired interval. In the following sections, we turn to the discussion of the discrete (empirical) case, where the data set consists of a finite collection of numbers rather than a continuous distribution.

\subsection*{6.6 Scaling by $b$ and the representation $g(s,b)=R(\beta-s)-R(\beta)$}
\label{subsec:scaling-R}

\paragraph*{Why this matters.}
Working with
\[
g(s,b)=\int f(z/b)\,M(s,z)\,\frac{dz}{b},\qquad b=10^\beta,\ \beta=\{\log_{10} b\},
\]
gives a universal way to compare “mantissa profiles” across scales and units. This is crucial when data are normalized by an arbitrary bound $B$ or naturally appear in a scaled form $X/B$ (e.g., on $[0,B]$).

\begin{itemize}
\item \textbf{Comparability across scales and units.}
The form $f(z/b)/b$ automatically accounts for stretching/compressing the axis, enabling fair comparison of datasets measured in different units or over different ranges.
\item \textbf{Diagnostics of leading-digit structure.}
The profile $g(s,b)$ averages the indicator $\{\!\log_{10}(bz)\!\}<s$, i.e., a modulus on the log scale. It supports quality control, anomaly detection, and DLSD-style analysis, including for samples truncated to $[0,B]$.
\item \textbf{Broad coverage of continuous laws.}
Many distributions (power laws on $[0,B]$, Weibull, normal, lognormal, gamma, etc.) naturally admit the scaling $f(z/b)/b$, so their mantissa profiles fit into a single framework.
\item \textbf{Operational recipe via a base slice.}
The representation $g(s,b)=R(\beta-s)-R(\beta)$ with $\beta=\{\log_{10} b\}$ implies that once the base slice $g_1(s)=g(s,1)$ is known, we can obtain $g(s,b)$ for \emph{any} $b>0$ immediately (see \eqref{eq:piecewise-g} and \eqref{eq:g-as-R}).
\end{itemize}

\paragraph{Setup and notation.}
Fix base $10$. For $t>0$ write $\{\log_{10} t\}\in[0,1)$ for the fractional part of $\log_{10} t$, and let $s\in[0,1]$. Define the window
\begin{equation}\label{eq:M-def}
M(s,z)\ :=\ -\,\Big\lfloor \{\log_{10} z\}-s\Big\rfloor
\qquad (z>0),
\end{equation}
so that $M(s,z)=\mathbf{1}_{\{\{\log_{10} z\}<s\}}$. Let $f$ be a density on $(0,\infty)$. We consider two equivalent normalizations:
\begin{align}
g(s,b)
&=\frac{1}{b}\int_{0}^{b} f\!\left(\frac{z}{b}\right) M(s,z)\,dz
=\int_{0}^{1} f(x)\, M(s,bx)\,dx, \label{eq:g-compact-support}\\[2mm]
g(s,b)
&=\frac{1}{b}\int_{0}^{\infty} f\!\left(\frac{z}{b}\right) M(s,z)\,dz
=\int_{0}^{\infty} f(x)\, M(s,bx)\,dx. \label{eq:g-infinite-support}
\end{align}
(Use $x=z/b$.) It is often convenient to write $b=10^{m+\beta}$ with
\[
m=\lfloor \log_{10} b\rfloor\in\mathbb{Z},\qquad \beta=\{\log_{10} b\}\in[0,1).
\]
Then $z=y\,10^{m}$ and $y=10^{\beta}x$ make the scale separation explicit.

\paragraph{Step 1 (base slice).}
Define
\begin{equation}\label{eq:g1-def}
g_1(s)\ :=\ g(s,1)
=\int f(x)\, \mathbf{1}_{\{\{\log_{10} x\}<s\}}\,dx
=\sum_{k\in\mathbb{Z}}\;\int_{10^{k}}^{10^{k+s}} f(x)\,dx ,
\end{equation}
where the last equality uses $\{\log_{10} x\}<s \iff x\in\bigcup_{k\in\mathbb{Z}}[10^k,\,10^{k+s})$.

\paragraph{Step 2 (decomposing $b$).}
Write $b=10^{m+\beta}$. For any $x>0$,
\[
\{\log_{10}(bx)\}=\{\log_{10} b+\log_{10} x\}=\{\beta+\{\log_{10} x\}\}.
\]
Let $U:=\{\log_{10} X\}\in[0,1)$ for $X\sim f$. Then
\begin{equation}\label{eq:g-as-shifted-CDF}
g(s,b)=\mathbb{P}\bigl(\{\beta+U\}<s\bigr).
\end{equation}

\paragraph{Step 3 (piecewise formula for $g(s,b)$ via $g_1$).}
Let $F_U(s):=\mathbb{P}(U<s)=g_1(s)$ on $[0,1]$ (use right/left limits if needed). From \eqref{eq:g-as-shifted-CDF} we obtain the circular-shift identities
\begin{equation}\label{eq:piecewise-g}
g(s,b)=
\begin{cases}
g_1(1-\beta+s)-g_1(1-\beta), & 0\le s\le \beta,\\[4pt]
1+g_1(s-\beta)-g_1(1-\beta), & \beta<s\le 1,
\end{cases}
\end{equation}
so that $g(0,b)=0$, $g(1,b)=1$, and $g(s,1)=g_1(s)$.

\paragraph{Step 4 (the $R$-representation and recovery of $R$).}
Define
\begin{equation}\label{eq:R-from-g1}
R(u)\ :=\ g_1\!\bigl(1-\{u\}\bigr)-1-\lfloor u\rfloor .
\end{equation}
Then $R(u+1)=R(u)-1$ (so $H(u):=R(u)+u$ is $1$-periodic), and for every $s\in[0,1]$ and $\beta=\{\log_{10} b\}$,
\begin{equation}\label{eq:g-as-R}
\boxed{\,g(s,b)=R(\beta-s)-R(\beta).\,}
\end{equation}
\emph{Proof sketch.} Write $R(u)=-u+H(\{u\})$ with $H$ $1$-periodic. Choosing $H(0)=0$ and solving $g_1(s)=R(-s)-R(0)$ yields $H(u)=g_1(1-u)-(1-u)$; substituting gives \eqref{eq:R-from-g1} and then \eqref{eq:g-as-R}. Uniqueness holds up to an additive constant, which cancels in the difference.

\paragraph{Computational note.}
For any density $f$ on $(0,\infty)$,
\[
g_1(s)=\sum_{k\in\mathbb{Z}}\int_{10^{k}}^{10^{k+s}} f(x)\,dx.
\]
Numerically: (i) tabulate $g_1$ on a grid of $s\in[0,1]$ (truncate the sum where $f$ is negligible); (ii) build $R$ via \eqref{eq:R-from-g1}; (iii) evaluate $g(s,b)$ for any $b>0$ using \eqref{eq:piecewise-g} or \eqref{eq:g-as-R}.

\paragraph{Example (uniform density on $[0,b)$).}\label{ex:uniform01}
Let $f(x)=\mathbf{1}_{(0,1)}(x)$. Then
\[
g_1(s)=\int_{0}^{1}\mathbf{1}_{\{\{\log_{10}x\}<s\}}\,dx
=\sum_{n=0}^{\infty}\int_{10^{-n-1}}^{10^{-n}}\mathbf{1}_{\{\{\log_{10}x\}<s\}}\,dx
=\sum_{n=0}^{\infty}\bigl(10^{-n-1+s}-10^{-n-1}\bigr)
=\frac{10^{s}-1}{9}.
\]
Thus, with $b=10^{m+\beta}$ and $\beta=\{\log_{10}b\}$, the piecewise form \eqref{eq:piecewise-g} gives
\[
g(s,b)=
\begin{cases}
\dfrac{10}{9}\,10^{-\beta}\bigl(10^s-1\bigr), & 0\le s\le \beta,\\[8pt]
1+\dfrac{10^{-\beta}}{9}\bigl(10^s-10\bigr), & \beta<s\le 1,
\end{cases}
\]
and one convenient choice of $R$ is
\[
R(u)=\dfrac{10^{\,1-\{u\}}-10}{9}-\lfloor u\rfloor .
\]
\emph{Remark.} This example is \textbf{not} an appeal to “Benford’s law”; the equality
$g_1(s)=(10^s-1)/9$ arises from uniform $x\!\in\!(0,1)$ and the geometric partition of $(0,1)$
into decades. It merely coincides with the base-10 Benford mantissa CDF.


\section*{VII Discrete Distributions and Empirical Digital Profiles}
\addcontentsline{toc}{section}{VII. Discrete Distributions and Empirical Digital Profiles}

In this section, we develop the discrete analogue of the density and distribution functions as applied to digital statistics. The motivation comes from the analysis of finite datasets, where the standard notion of probability density is no longer applicable, and empirical methods must be used.

\subsection*{7.1 Empirical Density for a Finite Set}

Suppose we are given a finite set of $N$ positive real numbers $\{a_t\}_{t=1}^N$.  
Before analysis, all leading zeros and decimal points are removed, and only strictly positive numbers are retained, ensuring every number is properly formatted for digital analysis.

The empirical density function is defined as:
\[
f(x) = \frac{1}{N} \sum_{t=1}^{N} \delta(x - a_t)
\]
where $\delta(x - a_t)$ is the Dirac delta function centered at $a_t$.  
In the discrete context, this means $f(x)$ is nonzero only at those $x$ which coincide with one of the $a_t$, and its value is the normalized count (frequency) of $x$ in the dataset.

\subsection*{7.2 Empirical Distribution Function}

The corresponding empirical distribution function is:
\[
F(x) = \frac{1}{N} \sum_{t=1}^{N} \mathbf{1}_{\{a_t \leq x\}}
\]
where $\mathbf{1}_{\{a_t \leq x\}}$ is the indicator function, equal to 1 if $a_t \leq x$, and 0 otherwise.

\subsection*{7.3 Application to Digital Statistics}

This approach allows us to empirically analyze the distribution of leading digits in any finite set of positive real numbers, regardless of their origin or structure.  
In particular, it provides a foundation for the study of deviations from classical digital laws (such as Benford's Law) in real-world datasets.

\textbf{Remark:}  
The empirical approach is particularly well suited for analyzing digital statistics in practical data, where sample sizes are finite and distributions may be highly irregular or even multimodal.

\subsection*{7.4 The Empirical Leading Digit Density: $G$-function and $V$-function Approaches}

There are two equivalent approaches to computing the empirical leading digit frequency (density) $\rho(k)$ in a finite sample $\{a_i\}_{i=1}^N$:

\textbf{1. Through the $G$-function:}
\[
\rho(k) = \lfloor \log_{10}(k+1) \rfloor - \lfloor \log_{10}(k) \rfloor + G(\{\log_{10}(k+1)\}) - G(\{\log_{10}(k)\})
\]
where $G(s)$ is the leading digit profile function defined earlier, and $\{x\}$ denotes the fractional part.

\textbf{2. Through the $V$-function (direct counting):}
\[
V(k, x) = \left\lfloor \log_{10}\left(\frac{x}{k}\right) \right\rfloor - \left\lfloor \log_{10}\left(\frac{x}{k+1}\right) \right\rfloor
\]
\[
\rho(k) = \frac{1}{N} \sum_{i=1}^{N} V(k, a_i)
\]
The $V$-function is $1$ if $k$ is the leading digit of $x$, and $0$ otherwise. Thus, this sum simply counts the fraction of numbers in the sample that begin with digit $k$.

\textbf{Remark:}  
The $G$-function and $V$-function approaches are consistent; in the discrete, empirical setting, the direct counting method via $V$ is the most natural and computationally straightforward.


\subsection*{7.5 Single Number Example: Canonicalization and Infinite–Digit View}

A notable strength of our empirical digital profile method is that it applies even to the smallest datasets, including the case of a single number. However, for correct digital analysis, we must exclude trivial representations where the number is written as $1$, $10$, $100$, $1000$, and so on, since all such forms correspond to the same significant value with varying numbers of trailing zeros.

\subsection*{7.5.1 Single Number: Blocks for All $m$ (one datum, infinitely many blocks)}

\paragraph{Digits of a single number.}
Write any $x>0$ in scientific form $x=10^{n(x)}\,m(x)$ with $n(x)\in\mathbb{Z}$
and $m(x)\in[1,10)$.  Let the mantissa have the (unique) digit expansion
\[
m(x)=d_0.d_1d_2d_3\ldots,\qquad d_0\in\{1,\dots,9\},\ \ d_j\in\{0,\dots,9\}.
\]
Removing the decimal point gives the \emph{infinite significant–digit string}
\[
\boldsymbol{\sigma}(x)=d_0d_1d_2d_3\ldots
\]
— even a single number carries infinitely many significant digits.

\paragraph{Leading $m$-digit block of a single number.}
For each $m\ge1$ define the leading $m$-digit block
\[
B_m(x)\;=\;\Big\lfloor 10^{\,m-1+\{\log_{10}x\}}\Big\rfloor
\;=\;\sum_{j=0}^{m-1} d_j\,10^{m-1-j}\ \in\ \{10^{m-1},\dots,10^m-1\}.
\]
If the dataset is the \emph{single} observation $\{x\}$, then the empirical
block distribution of order $m$ is the point mass
\[
\rho_m(k)\;=\;\mathbf{1}\!\bigl(k=B_m(x)\bigr) ,\qquad
k=10^{m-1},\dots,10^m-1.
\]
In words: for that one number, the block $B_m(x)$ has probability $1$ and all
other $m$-blocks have probability $0$.

\paragraph{Examples (one number, many blocks).}
\begin{itemize}
\item $x=1=1.000\ldots$ has digits $1,0,0,\ldots$ hence
      \[
      B_1(1)=1,\quad B_2(1)=10,\quad B_3(1)=100,\ \ldots,\ B_m(1)=10^{m-1}.
      \]
\item $x=3=3.000\ldots$ has digits $3,0,0,\ldots$ hence
      \[
      B_1(3)=3,\quad B_2(3)=30,\quad B_3(3)=300,\ \ldots,\ B_m(3)=3\cdot10^{m-1}.
      \]
\item $x=\pi\approx3.14159\ldots$ has digits $3,1,4,1,5,9,\ldots$ hence
      \[
      B_1(\pi)=3,\quad B_2(\pi)=31,\quad B_3(\pi)=314,\quad B_4(\pi)=3141,\ \ldots
      \]
\end{itemize}

\paragraph{No duplication by decimal scaling.}
Numbers $x$ and $10^{k}x$ ($k\in\mathbb{Z}$) have the same digit string
$\boldsymbol{\sigma}$ and thus the same blocks $B_m(\cdot)$; we therefore do
\emph{not} count $x,10x,10^2x,\ldots$ as separate observations when the dataset
is a single number.

\textbf{Conclusion:}  
This illustrates the complete generality of the empirical approach: it yields exact, arithmetic results for all leading digit blocks, not just the first digit, making it a powerful tool for digital analysis at any desired scale.


\subsection*{7.6 Arithmetic Sequences and the \(G(s)\) Function}

Consider the arithmetic sequence
\[
a_i = \alpha i + \beta,
\]
where \(\alpha > 0\), \(\beta \geq 0\), and both \(\alpha\) and \(\beta\) are real numbers. The index \(i\) runs from \(1\) to \(N\).

We define the function \(G(s)\) as the sum
\[
G(s) =\frac{1}{N} \sum_{i=1}^{N} M(s, a_i),
\]
where \(M(s, a_i)\) is a function of \(s\) and \(a_i\). Here, \(a_i\) takes the form specified above.

We will study the behavior of the function \(G(s)\) for real associative values of the parameters and analyze how its properties depend on the sequence parameters \(\alpha\), \(\beta\), and \(N\).

We define the function \( J(s) \) as follows:
\[
J(s) = \frac{1}{N} \int_{z=1}^{N} -\left\lfloor \left\{ \log_{10}(\alpha z + \beta) - s \right\} \right\rfloor \, dz
\]
where \( \alpha > 0 \), \( \beta \geq 0 \), and \( N \) is the upper limit of the variable \( z \).

Now, let us make a change of variables. Set \( y > z \), with \( z = \frac{y - \beta}{\alpha} \). In these terms, the integral for \( J(s) \) can be rewritten as:
\[
J(s) = \frac{1}{N} \int_{y = \alpha + \beta}^{\alpha N + \beta} -\left\lfloor \left\{ \log_{10}(y) - s \right\} \right\rfloor \frac{dy}{\alpha}
\]

Here, we performed the substitution from \( z \) to \( y \), setting the stage for further analysis.

With the change of variables and normalization, the function \( J(s) \) can be written as
\[
J(s) = \frac{1}{b - \beta/\nu} \int_{x = (\alpha + \beta)/\nu}^{b} -\left\lfloor \left\{ \log_{10}(x) - s \right\} \right\rfloor \, dx
\]
where
\[
\nu = 10^{\left\lfloor \log_{10}(\alpha N + \beta) \right\rfloor}, \quad b = \frac{10^{\log_{10}(\alpha N + \beta)}}{\nu}
\]

In the limit as \( N \to \infty \), the lower bound becomes negligible, and the formula simplifies to
\[
J(s) = \frac{1}{b} \int_{x=0}^{b} -\left\lfloor \left\{ \log_{10}(x) - s \right\} \right\rfloor \, dx
\]

\[
J(s) = \frac{10 \cdot 10^{-\{ -s + \{\log_{10} b\} \}}}{9}
      - \frac{10 \cdot 10^{-\{\log_{10} b\}}}{9}
      - \left\lfloor - s + \{\log_{10} b\} \right\rfloor
\]
where \( \{x\} = x - \lfloor x \rfloor \) denotes the fractional part of \( x \).
The function \( G(s) \) is originally defined as a sum, while \( J(s) \) is formulated as an integral. We argue that, for large \( N \), the integral provides an excellent approximation to the sum.

The key reason is that the summand in \( G(s) \) changes slowly with respect to its index, especially for large \( N \). Since the function is essentially logarithmic, its variation over a single interval is very small, and the difference between the discrete sum and the continuous integral becomes negligible as \( N \) increases. Thus, replacing the sum by the integral is justified; the resulting error is of order \( 1/N \) and vanishes in the limit \( N \to \infty \).

Additionally, the initial partition or lower bound of the summation or integral is unimportant in the limit. When dividing by \( N \), any fixed number of initial terms or the contribution from the lower limit disappears in the large \( N \) limit. In other words, the influence of the starting point is suppressed by the normalization, and only the main body of the sum or integral determines the result as \( N \) becomes very large.
\begin{center}
\begin{minipage}{0.85\textwidth}
\itshape
To conclude, our results demonstrate that the function $\rho(k)$ for an arithmetic sequence exhibits a striking periodic behavior, oscillating between its maximum and minimum values as dictated by the decimal structure. While numerical methods can illustrate these oscillations and confirm the bounds, the analytical formulas derived here provide an exact description valid for any $n$, whether or not the interval corresponds to a complete decade. This highlights not only the power of the analytical approach but also the rich structure underlying the distribution of leading digits. The combination of rigorous analysis and numerical evidence offers a comprehensive understanding of the phenomenon, paving the way for further exploration using both methods.
\end{minipage}
\end{center}
\bigskip

\subsection*{7.7 From Discrete Sums to Integral Approximations: Asymptotic Analysis of Digital Profiles}

The study of leading digit distributions in empirical datasets naturally begins with discrete sums. Suppose we have a finite sample $\{x_1, x_2, \ldots, x_N\}$, and let $M(s, x_i)$ denote an indicator or cumulative function associated with the leading digit profile (for example, $M(s, x) = 1$ if the leading digit of $x$ is less than $s$, and $0$ otherwise). The empirical digital profile is then defined as
\[
G(s) = \frac{1}{N} \sum_{i=1}^N M(s, x_i).
\]

\vspace{0.5em}
\textbf{Integral Approximation.}  
When the sample size $N$ is large and the data points $x_i$ can be parametrized smoothly (for example, $x_i = \varphi(i)$ for some monotonic function $\varphi$), it is often fruitful to approximate the discrete sum by a continuous integral:
\[
G(s) \approx \frac{1}{N} \int_{z=1}^{N} M(s, \varphi(z))\, dz.
\]
This substitution replaces the sum over discrete indices $i$ by an integral over a continuous variable $z$, which can often be analyzed with the powerful tools of calculus.

\vspace{0.5em}
\textbf{Change of Variables.}  
To further simplify the analysis, we may introduce a change of variables $y = \varphi(z)$, assuming $\varphi$ is invertible. The inverse function $\psi(y)$ satisfies $z = \psi(y)$, and $dz = \psi'(y)\, dy$. Substituting into the integral, we obtain
\[
G(s) \approx \int_{y = \varphi(1)}^{\varphi(N)} \psi'(y)\, M(s, y)\, dy.
\]
In this formulation, the function $\psi'(y)$ acts as an effective "density" for $y$ values, reflecting how the discrete points are distributed over the continuous range.

\vspace{0.5em}
\textbf{Interpretation and Limitations.}  
While this integral approximation is not exact in general—especially for small $N$ or highly irregular $x_i$—it often captures the asymptotic behavior of $G(s)$ as $N$ grows large. In many cases, it provides valuable intuition and a bridge between empirical (discrete) statistics and analytic (continuous) theory.

\vspace{0.5em}
\textit{Example:}  
If $x_i = i^\alpha$ for some $\alpha > 0$, then $\psi(y) = y^{1/\alpha}$ and $\psi'(y) = \frac{1}{\alpha} y^{1/\alpha - 1}$. The integral approximation becomes:
\[
G(s) \approx \int_{y = 1}^{N^\alpha} \frac{1}{\alpha} y^{1/\alpha - 1} M(s, y)\, dy.
\]

\vspace{0.5em}
\textit{Remark on normalization:}  
If the upper bound $\varphi(N)$ does not correspond to an exact decade (e.g., $\varphi(N) \neq 10^k$), one can define $\nu = 10^{\lfloor \log_{10} \varphi(N) \rfloor}$ as the nearest lower decade for normalization.

\vspace{0.5em}
In summary, this discrete-to-continuous transition provides a powerful analytic tool for understanding the asymptotic properties of digital distributions, though it must be used with care for finite or irregular datasets.

\textbf{Rescaling and Further Change of Variables.}

To analyze the asymptotic and normalized behavior of the distribution, it is often helpful to introduce a rescaling based on the characteristic scale $\nu$, defined as
\[
\nu = 10^{\lfloor \log_{10} \varphi(N) \rfloor}
\]
so that $\nu$ is the largest integer power of 10 not exceeding $\varphi(N)$. We can then make a further change of variable, setting $y = x \nu$, where $x$ runs over a normalized interval.

Let $b = \varphi(N)/\nu$ and $a = \varphi(1)/\nu$ so that $x \in [a, b]$ and $y \in [\varphi(1), \varphi(N)]$. Let $\psi_1(y) = \psi(y)$ for notational clarity. The cumulative profile in the rescaled variables becomes
\[
G(s) \approx \int_{y = \varphi(1)}^{\varphi(N)} \psi_1(y) M(s, y)\, dy
= \nu \int_{x = a}^{b} \psi_1(x\nu) M(s, x\nu) dx
\]

Normalizing by $N \nu$ and dividing by the effective width $b$ gives
\[
\lim_{N \to \infty} \frac{1}{N \nu} \int_{x = a}^{b} \psi_1(x \nu) M(s, x \nu) dx = \frac{1}{b} \int_{x = a}^{b} f_\infty(x) M(s, x) dx
\]
where $f_\infty(x)$ denotes the limiting density as $N \to \infty$. This expression captures the normalized contribution of each subinterval in the large-$N$ limit.

\textit{Interpretation.}  
This rescaling is particularly useful when the upper endpoint $\varphi(N)$ does not fall exactly on a decade. By normalizing with respect to $\nu$ and focusing on the relative position $x$ within the final decade, we can meaningfully compare distributions with different sample sizes or endpoints. In the limit $N \to \infty$, the scaled density $\psi_1(x\nu)$ approaches a universal profile on $[a, b]$.

\vspace{0.5em}
\textit{Summary.}  
The two-step change of variable—from discrete sum to integral, and then to rescaled, normalized variables—reveals the underlying scaling structure and facilitates the asymptotic analysis of digital distributions for large finite samples. While not exact for all datasets, this method provides a robust framework for understanding the limiting behavior and normalization of empirical digital profiles.

\textbf{Remark on the Weyl Equidistribution Theorem.}

It is worth noting that the integral approximation described above is closely related to Weyl's equidistribution theorem~\cite{weyl1916}. If the sequence $\{\log_{10} \varphi(x)\}$ (the fractional parts of $\log_{10} \varphi(x)$) is distributed according to a probability density $f(t)$ on the interval $[0,1)$, then in the limit $N \to \infty$, the empirical digital profile satisfies
\[
G(s) = \int_0^s f(t)\, dt.
\]
This is because, by the law of large numbers and Weyl's theorem, the normalized sum over $M(s, \varphi(x))$ converges to the integral of $f$ over $[0, s]$.

In other words, as $N$ increases, the fraction of elements with $\{\log_{10} \varphi(x)\} < s$ approaches the probability mass of $f(t)$ on $[0, s]$. This provides a rigorous foundation for approximating discrete digital distributions by integrals, especially in the presence of equidistribution or known limiting densities.

\subsection*{7.8 2023 World Population: Digital Density and Empirical Profile}

\paragraph{Dataset.} We use the 2023 world population table ({\it Total Population} only) with 217 economies. Male/Female breakdowns are available but not used here.

\begin{figure}[H]
  \centering
  \includegraphics[width=0.72\linewidth]{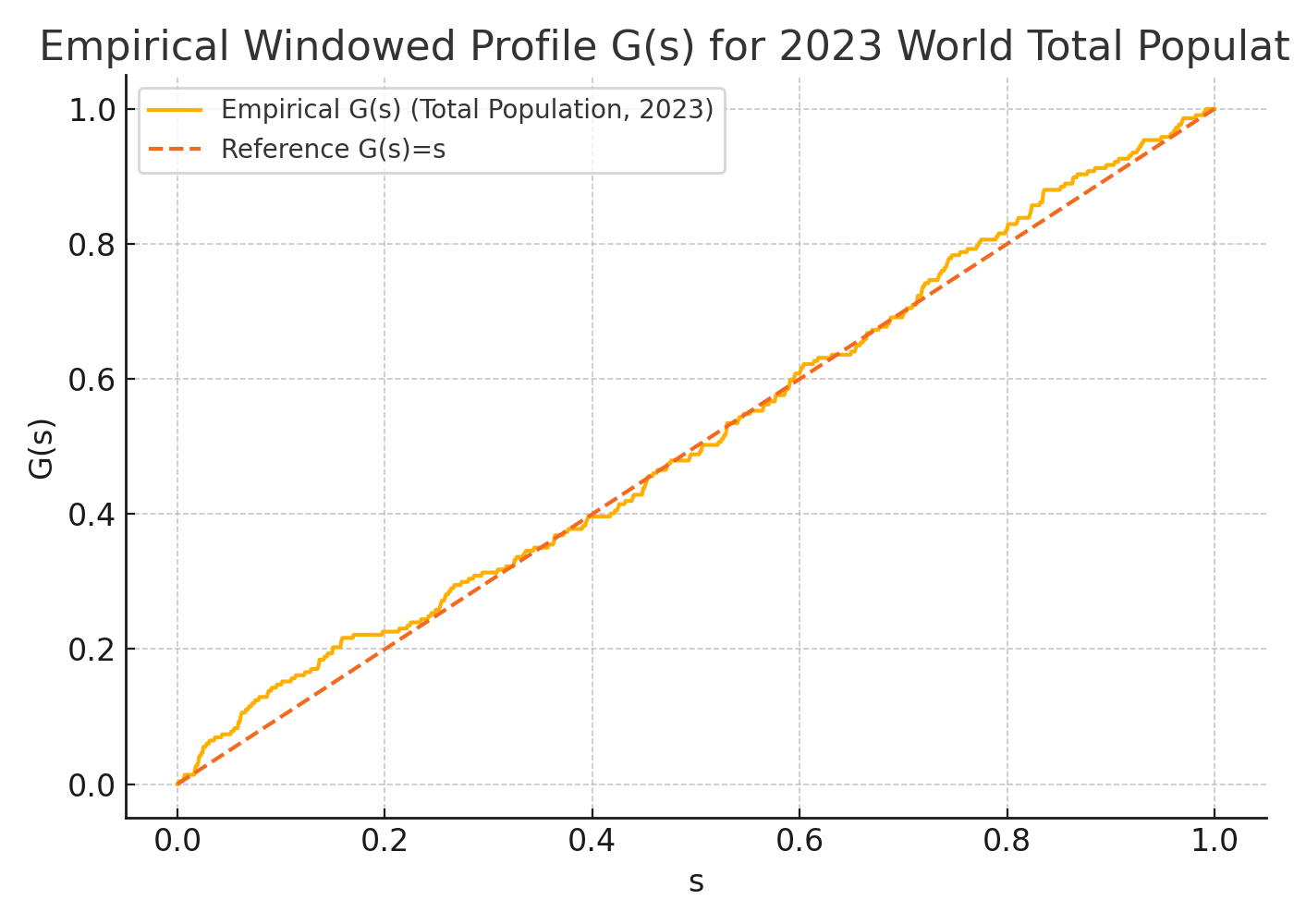}
  \caption{Empirical windowed profile $\widehat{G}(s)$ for 2023 world Total Population across countries (orange) with the reference line $G(s)=s$ (dashed).}
  \label{fig:pop2023_Gs}
\end{figure}

\paragraph{Method.} For each country we compute $s=\{\log_{10}(X)\}\in[0,1)$, the fractional part of the base-10 logarithm of the total population $X$. The empirical windowed profile is
\[
\widehat{G}(s)=\frac{1}{N}\sum_{i=1}^N \mathbf{1}\bigl(\{\log_{10}(X_i)\}\le s\bigr),
\]
plotted in Fig.~\ref{fig:pop2023_Gs}, together with the reference line $G(s)=s$.

\paragraph{First-digit proportions.} The observed first-digit frequencies for Total Population (countries, 2023) are shown in Table~\ref{tab:pop2023_first_digit}. As illustrative criteria, we report a Pearson chi-square against Benford's vector $\bigl(\log_{10}(1+1/d)\bigr)_{d=1}^9$ and the KL divergence $D_\mathrm{KL}(\widehat{p}\Vert p^\mathrm{Benford})$; these can be replaced by your DLSD-based criteria in the final draft.

\begin{table}[h]
  \centering
  \caption{First-digit distribution for 2023 Total Population across countries.}
  \label{tab:pop2023_first_digit}
  \begin{tabular}{r|rrrrrrrrr}
    \toprule
    Digit & 1 & 2 & 3 & 4 & 5 & 6 & 7 & 8 & 9 \\ \midrule
    Count & 68 & 36 & 30 & 16 & 25 & 16 & 8 & 9 & 9 \\
    Share & 0.313 & 0.166 & 0.138 & 0.074 & 0.115 & 0.074 & 0.037 & 0.041 & 0.041 \\ \bottomrule
  \end{tabular}
\end{table}

\paragraph{Goodness-of-fit (illustrative).} For this dataset ($N=217$), the chi-square statistic vs. Benford's vector is $7.609$, and the KL divergence is $0.01738$. These figures are reported only for reference; in the DLSD framework we will instead estimate your two-parameter profile $G(s)=F_W(s;b,a)$ and report its parameters and confidence bands. 


\subsection*{Weibull approximation for country-level Total Population (2023)}

\paragraph{Model and estimation.}
We fit the two-parameter Weibull distribution to the sample $X_i>0$ of country-level \emph{Total Population}.
The parametrization is $f(x;a,b)=\dfrac{a}{b}\left(\dfrac{x}{b}\right)^{a-1}\exp\!\left[-\left(\dfrac{x}{b}\right)^a\right]$,
with shape $a$ and scale $b$. Maximum likelihood yields
$\hat a=0.467$ and $\hat b=1.37467e+07$ (sample size $N=217$).

\paragraph{Goodness of fit.}
We report several diagnostics:
(i) Kolmogorov--Smirnov statistic with parametric bootstrap ($B=300$) gives $D=0.0454$ with $p\approx0.355$;
(ii) Pearson chi-square on $m=10$ equiprobable bins with parametric bootstrap ($B=400$) gives $\chi^2=5.72$ with $p\approx0.621$;
(iii) log-likelihood $\ell(\hat a,\hat b)=-3813.4$, AIC=7630.8, BIC=7637.5.

\emph{Interpretation.} The small KS distance ($D=0.0454$) together with a large parametric-bootstrap $p$-value ($p\approx0.355$) indicates that the fitted Weibull profile is statistically indistinguishable from the empirical $G(s)$ at conventional levels, and the Pearson test on equiprobable bins likewise shows no lack-of-fit ($\chi^2=5.72$, $p\approx0.621$). 
The penalized-likelihood scores (AIC$=7630.8$, BIC$=7637.5$) are consistent with a parsimonious two-parameter specification and do not signal overfitting, reinforcing that the Weibull approximation is adequate for these data.

\begin{figure}[H]
  \centering
  \includegraphics[width=0.78\linewidth]{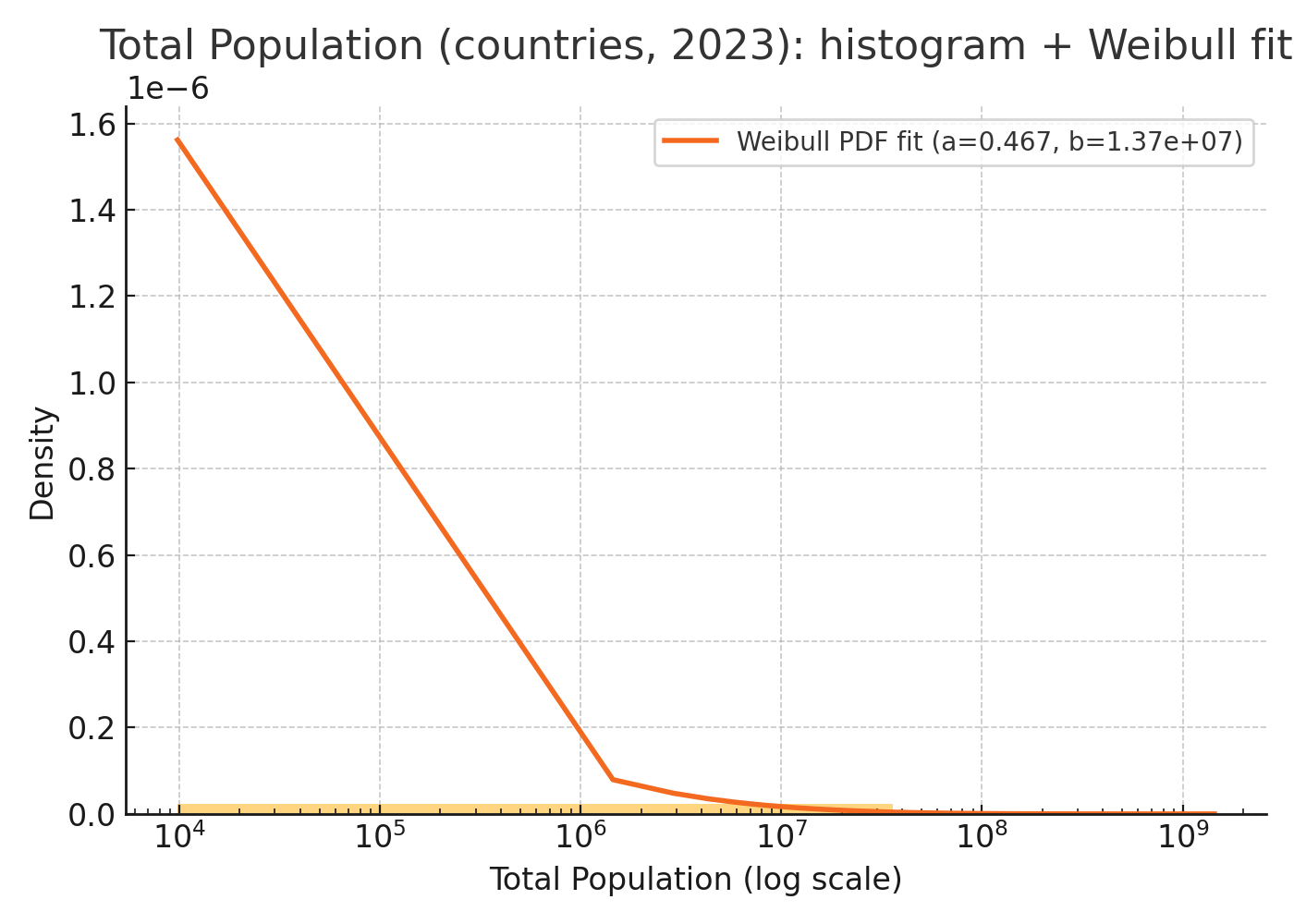}
  \caption{Histogram of Total Population (countries, 2023) on a log-$x$ axis with the fitted Weibull pdf overlay
  ($\hat a=0.467$, $\hat b=1.37467e+07$).}
  \label{fig:weibull_pop2023_fit}
\end{figure}

\begin{figure}[H]
  \centering
  \includegraphics[width=0.62\linewidth]{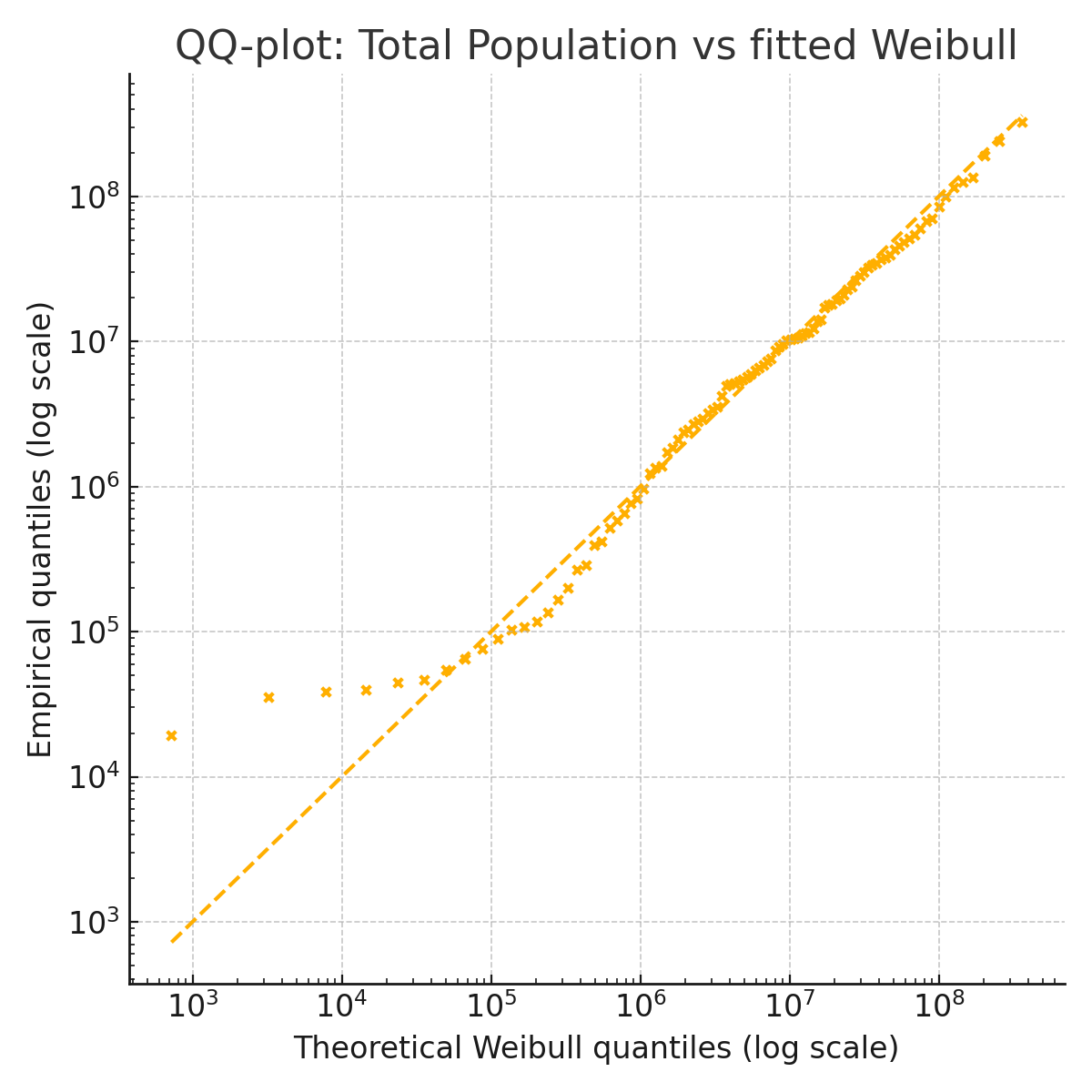}
  \caption{QQ-plot (log--log axes) comparing empirical quantiles of Total Population against the fitted Weibull quantiles.}
  \label{fig:weibull_pop2023_qq}
\end{figure}

\paragraph{Interpretation.}
For $a=1$ (the exponential case), the profile $G(s)$ almost coincides with the line $s$, reflecting that the distribution of phases $\{\log_{10}X\}$ is close to uniform.
For $a=0.5$, the deviation from $s$ is small and has a form consistent with heavier tails (a greater share of mass at small $x$).
For $a=3$, the deformation becomes pronounced and S-shaped: for small $s$ the profile grows faster than the straight line and then slows, which is associated with the narrower shape of the Weibull density and the concentration of mass near a characteristic scale.


\section*{VIII Critical Discussion}
\addcontentsline{toc}{section}{VIII Critical Discussion}
\noindent\makebox[\linewidth]{\rule{\textwidth}{0.4pt}}

This chapter offers a comprehensive critical analysis of digit laws, with a particular emphasis on Benford’s Law and the broader class of digit distributions. The discussion below synthesizes recent theoretical insights with foundational observations and highlights both the limitations and the universality claims often made in the literature.

\textbf{1. On Leading Versus First Digits.} \\
A key conceptual distinction must be drawn between “leading digit” and “first digit.” While “first digit” refers strictly to the initial nonzero digit of a number, “leading digit” can—depending on context—refer to a sequence of significant digits at the beginning of a number. In much of the Benford literature, “leading digit” is often treated as synonymous with “first digit.” Our approach clarifies that all leading digits, not just the first, are structurally interconnected by the positional nature of number systems. Understanding this connection is vital for a truly rigorous analysis of digit laws.

\textbf{2. The Algebraic—not Universal—Nature of Benford’s Law.} \\
Despite its frequent portrayal as a universal statistical law, Benford’s Law is fundamentally algebraic, rooted in the logarithmic and positional properties of number representations. Its apparent universality is a reflection of the decimal system, not an intrinsic law of nature.

\textbf{3. Critique of Mixture and Invariance Arguments.} \\
The claim that “random mixtures” of distributions inevitably produce Benford’s Law is circular: any mixture can be tailored to fit a target digit law, and such mixtures rarely arise spontaneously in practice. True universality is therefore not established by these arguments.

\textbf{4. The Interdependence of All Significant Digits.} \\
Focusing only on leading digits is an artificial limitation. All significant digits are mathematically linked, and a full understanding requires analyzing their joint distribution. Any separation between digits is a matter of historical convenience, not mathematical necessity.

\textbf{5. Structural and Deterministic Explanations.} \\
We argue for a shift from probabilistic “miracle” explanations toward explicit, structural, and combinatorial derivations of digit distributions. Many digit laws, including Benford’s, can be shown to arise naturally from the algebraic properties of sequences or products—no randomness is required.

\textbf{6. The Limits of “Naturalness” and “Universality.”} \\
There is no privileged or “natural” process that always yields Benford-type distributions. Many empirical datasets, especially those that are finite, bounded, or human-constructed, do not exhibit Benford’s Law. The observed “universality” is largely an artifact of particular mathematical constructions and dataset selections.

\textbf{7. The Failure of Universality for Small Samples.} \\
Benford’s Law and related digit laws do not generally hold for small or degenerate datasets. In these cases, digit frequencies are determined by the explicit structure of the data, not by any universal law.

\textbf{8. Logical Consistency and Mathematical “Stitching.”} \\
There are strict mathematical constraints linking the distributions of leading digits and those of longer digit blocks. Disregarding these links can lead to inconsistent “laws” that cannot be realized by any actual data.

\textbf{9. A New Program: Structural and Algorithmic Theory.} \\
We advocate for a new, fully structural theory of digit laws, based on explicit equations and combinatorics rather than probabilistic mixtures. This theory provides analytic tools for both theoretical and practical analysis, demystifies the phenomenon, and applies equally well to small samples as to large ones.

\vspace{0.5em}
In conclusion, the study of digit laws is enriched by recognizing the deep structural relationships among all significant digits, understanding the limitations of existing universality claims, and pursuing explicit analytic descriptions. Our approach aims to clarify these relationships and lay the foundation for a more precise and universally applicable theory of digit distributions.

\subsection*{8.1 The Universal Function \(G(s)\) and Interval Filling}

A fundamental observation is that, while the first (leading) digit can be seen as a discrete point—corresponding to a specific position between, say, 1 and 2—the inclusion of all subsequent digits fills out the entire interval. For example, if the leading digit marks a single point within the range from one to two, then the numbers with that leading digit (such as 10, 11, 12, etc.) densely populate the subinterval between those two points.

As we consider additional digits, each successive digit acts to further subdivide and fill in the interval. Ultimately, the collective contribution of all digits leads to a dense covering of the entire curve, so that what begins as a discrete set of points (the nine possible leading digits) becomes, in the limit, a continuous distribution spanning the full range from 0 to 1.

This property underlines the fact that the structure we study is not limited to a set of nine discrete points, but extends to a complete filling of the interval—a “curve” concretely determined by the arrangement and accumulation of all significant digits in the number system. Formally proving this property may require additional mathematical development, but the phenomenon is intuitively clear: the process of incorporating further digits results in an increasingly fine subdivision, eventually covering the entire line from zero to one.

\begin{remark}
A particularly clear and geometric criterion for the presence or absence of “Benford-like” properties in a dataset is provided by the $G(s)$ function. The closer $G(s)$ lies to the diagonal from $(0,0)$ to $(1,1)$, the more the digit frequencies resemble the logarithmic law. Significant deviations from this diagonal signal the absence of Benford-type behavior. In most empirical and mathematical distributions, the $G(s)$ function is notably non-linear, which visually and quantitatively illustrates why traditional digit laws are far from universal.
\end{remark}

\begin{remark}
A fascinating feature of leading digit distributions emerges when one examines the sequence of fractional parts $\left\{ \log_{10} k \right\}$ as $k$ ranges over successive decades. For $k = 1,\dots,9$, these points occupy exactly nine distinct locations in $[0,1)$; for $k = 10,\dots,99$, the new points interleave between those of the previous decade, and so on. Each successive decade adds a layer of finer subdivision, filling the unit interval with increasingly dense and nested sets of points. This iterative, scale-dependent process naturally leads to {\bfseries\itshape fractal-like}   
 , self-similar structure in the distribution of $\left\{ \log_{10} k \right\}$ as $k$ increases. Such nested arrangements provide an intuitive geometric explanation for the gradual, structured approach to the limiting distribution, and underscore the deep connections between digital laws, number theory, and fractal geometry. Illustrative plots of $\left\{ \log_{10} k \right\}$ for growing $k$ (e.g., for $k$ up to $10$, $100$, $1000$) make this structure immediately visible.
\end{remark}

\begin{remark}
A fundamental difference between the traditional “Benford paradigm” and the deterministic framework presented here lies in the direction of constraint. In Benford's Law, the frequencies of the first nine digits (1–9) uniquely determine the entire digit profile: all subsequent digit probabilities are forced to lie precisely on the logarithmic Benford curve, regardless of the specifics of the data. In contrast, within the DLSD framework, the arrangement of significant digits beyond the first can be essentially arbitrary—each additional choice further shapes the global curve, rather than being dictated by a universal law. Thus, in the deterministic setting, the observed profile is a consequence of the underlying sequence, not an imposed template, and the digit distribution encodes the structural “fingerprint” of the data rather than conforming to a predetermined law.
\end{remark}

\begin{remark}
This contrast can be illustrated even more vividly with an artificial example. Consider a distribution where the first nine significant digits are arranged to perfectly match the logarithmic probabilities typically associated with the Benford's Law. In the traditional setting, this would force the entire digit law to coincide with the canonical curve, and every point in the digit profile would be dictated by this initial choice. 

However, in the deterministic DLSD framework, the points corresponding to the first nine digits (from 1 to 9) remain fixed, but all additional points (for blocks or numbers beyond the initial set, such as $k=10$, $k=21$, etc.) are free to be chosen in essentially any configuration. These further choices dynamically generate the shape of the global curve, rather than being consequences of a universal template. Thus, the overall profile is not prescribed from above; rather, it emerges “from below,” as the collective outcome of specific, possibly non-logarithmic, placements of additional digits. 

This principle not only underlines the philosophical difference between the traditional and deterministic approaches, but also explains how a wide variety of leading digit distributions can arise, all sharing identical “first digit” statistics yet exhibiting diverse structures elsewhere in the digital profile.
\end{remark}

%
%
%

\subsection*{8.2 Asymptotic Paradox: Structure Preserved in the Infinite Limit}

It is widely believed—even in classic works by Benford himself—that as digit blocks become large, all empirical digit distributions must converge to Benford’s law in the infinite limit. However, this assertion is incorrect. Our analysis establishes, for the first time, the true structure of digit–block asymptotics and reveals that \textbf{the limiting frequency of digit blocks retains an explicit analytic memory of the generating process—even as $k\to\infty$}.

\medskip
\noindent
\textbf{\emph{General asymptotic formula.}} For any analytic generating profile $G(s)$, with $s=\{\log_{10}k\}$, the correct asymptotic for interior $k$ is
\[
\rho(k)\sim\frac{G'(s)}{k\,\ln 10},
\]
where $G(s)$ encodes the structural origin of the process generating the data.

\medskip
\noindent
\textbf{\emph{Key examples.}}
\begin{itemize}
\item \emph{Benford case:} For $G(s)=s$, $G'(s)=1$, so
\[
\rho(k)\sim\frac{1}{k\,\ln 10},
\]
which recovers the classic Benford law in the infinite limit.

\item \emph{Linear–uniform window:} For $G(s)=\dfrac{10^s-1}{9}$, $G'(s)=\dfrac{\ln 10}{9}10^{s}$, and with $n=\lfloor\log_{10}k\rfloor$,
\[
\rho(k)\sim\frac{1}{9\,10^{\,n}},
\]
i.e., the block probability is constant within each decade, not decaying as $1/k$.
\end{itemize}

\medskip
\noindent
\textbf{\emph{Implication.}} Contrary to the folklore, \textbf{the infinite limit does not erase the origin of the distribution}. Instead, the asymptotic digit law \emph{preserves a precise analytic “fingerprint” of its source}, even for arbitrarily large digit blocks. The celebrated “universality” of Benford’s law is merely a degenerate special case within a much broader space of possible asymptotic behaviors. This resolves a long-standing paradox and provides a deterministic, structural explanation for the diversity of observed digit distributions in empirical data.


\vspace{0.5em}
\noindent
\emph{Remark (Originality):} To our knowledge, this explicit and general asymptotic formula, and the recognition that structural memory persists at infinity, has not previously appeared in the literature. It is a principal original contribution of this work.


\section*{IX. Open Problems and Future Directions}
\addcontentsline{toc}{section}{IX. Open Problems and Future Directions}

The study of leading digit laws, as well as their deterministic and structural foundations, opens up a wide spectrum of unsolved problems and intriguing mathematical phenomena. While the present work has developed a modern deterministic framework and provided critical analysis of classical probabilistic approaches, it is clear that much remains to be discovered about the deeper structure and universal properties of digit distributions.

\subsection*{Potential Fractal Properties and Self-Similarity}

One particularly stimulating direction for future research lies in exploring the possible fractal or self-similar characteristics of leading digit sets and their associated mappings. When mapping blocks of digits onto the unit interval using the fractional part of the logarithm, a fascinating "filling" process occurs: while the set of values $\{\log_{10} k\}$ for single digits $k=1, \ldots, 9$ occupies only isolated points, the expansion to larger digit blocks—such as numbers from $10$ to $19$, $20$ to $29$, etc.—progressively fills the interval $[0,1)$ in a complex, seemingly hierarchical pattern.

It would be valuable to formalize and rigorously analyze the distributional and geometric properties of these sets. In particular, questions arise such as:
\begin{itemize}
    \item Does the set $\{\log_{10} n : n \in \mathbb{N}\}$ (modulo 1) possess a fractal or multifractal structure?
    \item How does the addition of more digits or larger blocks affect the density and uniformity of coverage of $[0,1)$?
    \item Can we define and compute the fractal dimension (e.g., Hausdorff or Minkowski) of such sets?
    \item Are there natural self-similarities or scaling laws governing the "gaps" and "clusters" within these sets?
\end{itemize}
Initial observations suggest that the recursive nature of digit expansion in positional systems may indeed encode self-similarity or even fractality, but a precise mathematical characterization remains an open challenge.

\subsection*{Further Unsolved Problems}

In addition to the fractal questions, several other compelling problems emerge:
\begin{itemize}
    \item \textbf{Analytic Structure of Digital Profiles.} How does the analytic behavior of the cumulative profile function $g(s)$ depend on the underlying digit block size and base? What are the implications for the universality (or non-universality) of Benford-type laws?
    \item \textbf{Generalization to Other Bases and Systems.} How do the results extend or change for different number bases, non-integer bases, or even non-positional systems? Are there analogs of Benford’s Law with analogous deterministic frameworks?
    \item \textbf{Limits of Deterministic Laws for Empirical Data.} What are the boundaries of applicability for deterministic models, especially in noisy or highly non-uniform empirical datasets? Under what conditions do random or probabilistic models regain explanatory power?
    \item \textbf{Connection to Equidistribution and Diophantine Approximation.} How deeply are leading digit distributions linked to classical results in uniform distribution mod 1 and Diophantine approximation? Can deeper connections to ergodic theory or the theory of normal numbers be established?
    \item \textbf{Computational Complexity and Algorithms.} What is the computational complexity of simulating or testing for specific digital profiles? Can efficient algorithms be developed for empirical testing or data validation in real-world settings?
\end{itemize}
\subsection*{Synthetic and Specialized Sequences: Toward a Broader Digital Phenomenology}

An exciting avenue for future research lies in the synthetic exploration of leading digit laws and digital phenomena for various nontrivial and structurally rich sequences—far beyond the traditional examples or random sampling from standard distributions. In particular, sequences such as primes, factorial numbers and they combinations, and other arithmetic or combinatorial constructions provide a rich landscape for discovering new and unexpected behaviors in digital laws.

These specialized sequences often exhibit digit distributions with distinctive features: for example, persistent block structures, non-logarithmic profiles, or fractal-like patterns in the distribution of significant digits. Unlike typical datasets arising from random processes, these deterministic or structured sequences may display highly non-generic behaviors, sometimes reflecting hidden algebraic or modular regularities.

Understanding the mechanisms that give rise to such digit patterns, as well as developing criteria for distinguishing structural digital phenomena from purely statistical artifacts, represents a major open problem for future research. In particular, it is of great interest to classify the range of possible leading digit distributions arising from deterministic recurrences, modular constructions, or algebraic constraints, and to explore the role of self-similarity, arithmetic “resonances,” and block structures in shaping these distributions.

A systematic investigation in this direction would not only broaden our theoretical understanding of digital laws, but also yield practical methods for the synthesis and detection of digital patterns in both natural and artificial datasets. The continued study of specialized sequences thus holds promise for revealing deeper connections between number theory, combinatorics, and the phenomenology of significant digits.

\subsection*{Digital Laws for Classical and Synthetic Sequences: Literature and Perspectives}

The digital properties of many classical sequences—such as primes, Fibonacci numbers, and factorials—have already been studied in depth. For instance, it is known that both the Fibonacci sequence and factorials asymptotically obey Benford's Law, while the sequence of prime numbers, although showing some digital irregularities, has also been analyzed in this context~\cite{diaconis1977,Kontorovich2014,WeissteinFibo}.

Thus, while the most prominent classical examples are well understood, many open questions remain for less-explored or deliberately constructed ("synthetic") sequences, such as those built via modular arithmetic, rare digital patterns, or other non-standard rules (for example, like the Ulam sequence, defined recursively by summing the two smallest distinct previous terms in only one way).  There is particular interest in studying whether these sequences exhibit non-classical digital behavior, possible fractal phenomena, or other novel features not captured by traditional Benford-type laws.

A comprehensive exploration of such synthetic digital structures—both theoretically and computationally—remains a promising direction for future research.


\subsection*{Concluding Remarks}

In conclusion, the landscape of leading digit laws and digital phenomena is far from being fully charted. The deterministic framework developed here provides a rigorous alternative to traditional probabilistic thinking, but at the same time, it exposes subtle geometric, analytic, and potentially fractal structures that warrant deeper investigation. Future research could lead to a more unified theory encompassing both deterministic and random digital phenomena, with applications ranging from theoretical mathematics to data science, cryptography, and beyond.


%
%
%
\section*{Epilogue: Synthesis}
\addcontentsline{toc}{section}{Epilogue: Synthesis}


\begin{quote}
\emph{``Not everything that counts can be counted,\\
and not everything that can be counted counts.''}\\
\hfill --- attributed to Albert Einstein
\end{quote}

\medskip

\noindent
After hundreds of pages tracing the hidden logic of numbers, distributions, and digital laws, one may wonder: does it all amount only to statistics and formulas, or is there a deeper message? The search for regularity amid apparent chaos, the skepticism toward “universal” laws, and the insistence on structure as the true signature of meaning—these themes have guided this book.

Mathematics, at its heart, is not the worship of chance or a blind faith in empirical patterns, but the art of discerning structure where others see only randomness. The greatest challenge, and perhaps the deepest satisfaction, is to separate what can truly be explained from what merely appears “universal” by accident or incomplete perspective.

To close, here is a brief acrostic—part reflection, part invitation—for all who continue this search.

\section*{Acrostic: \textit{STRUCTURE IS TRUTH}}

\begin{flushleft}
\textbf{S}ome kneel before the altar of pure chance,\\
\textbf{T}heir eyes shut tight in probability's dance.\\
\textbf{R}eal patterns --- brushed aside as sin,\\
\textbf{U}ntested claims they fold neatly in.\\
\textbf{C}ircles of logic, wide and self-made,\\
\textbf{T}he data conforms --- once it's betrayed.\\
\textbf{U}nseen by them, the digits sing,\\
\textbf{R}esponding to structure, not random string.\\
\textbf{E}ach claim of ``law'' dissolves in air ---\\[1em]
\textbf{I}t’s noise they praise, not truth laid bare.\\
\textbf{S}till, we persist, though fashion scorns,\\[1em]
\textbf{T}racing the paths where meaning forms.\\
\textbf{R}igorous steps, not mystic smoke,\\
\textbf{U}ndo the myths the crowd invokes.\\
\textbf{T}his is no trick, no hidden bluff ---\\
\textbf{H}ere, structure speaks. And that's enough.
\end{flushleft}

\medskip

\begin{flushright}
\emph{--- Thaddeus of Linchester (c.~1594, or perhaps a myth)}
\end{flushright}

\medskip

\noindent
\textit{Structure is not always visible at first glance. But it is always there for those who seek—not by habit, but by proof.}

\cleardoublepage
\addcontentsline{toc}{section}{References}   
\bibliographystyle{unsrt}  

\bibliography{bibs/cleaned_dlsd_refs}

\end{document}